\documentclass[lettersize,journal]{IEEEtran}
\usepackage{amsmath,amsfonts}
\usepackage{array}
\usepackage[caption=false,font=normalsize,labelfont=sf,textfont=sf]{subfig}
\usepackage{textcomp}
\usepackage{stfloats}
\usepackage{url}
\usepackage{verbatim}
\usepackage{graphicx}
\usepackage{cite}
\usepackage{CJKutf8}
\usepackage{newclude} 
\usepackage{hyperref}
\usepackage{amsmath} 

\newcommand{\Rmnum}[1]{\uppercase\expandafter{\romannumeral #1}}  

\usepackage[most]{tcolorbox}
\usepackage{color}

\usepackage{soul}
\let\lyx\relax 
\let\hl\relax 

\usepackage{algorithm}
\usepackage{algorithmic}
\usepackage{hhline}
\usepackage{multirow}
\usepackage[normalem]{ulem}
\useunder{\uline}{\ul}{}
\usepackage{graphicx}
\usepackage[export]{adjustbox}
\usepackage{arydshln}
\begin{document}
    \begin{CJK}{UTF8}{gbsn}
        \title{
UniPoll: A Unified Social Media Poll Generation Framework via Multi-Objective Optimization
}

\author{

Yixia Li$^{\dagger}$, Rong Xiang, Yanlin Song$^{\dagger}$, Jing Li

\thanks{This work was substantially supported by a grant from the Research Grants Council of the Hong Kong Special Administrative Region, China (Project No. PolyU/25200821), the Innovation and Technology Fund (Project No. PRP/047/22FX), the NSFC Young Scientists Fund (Project No.62006203), and the PolyU RC-DSAI (1-CE1E). \textit{(Corresponding author: Jing Li.)}}

\thanks{Rong Xiang and Jing Li are with the Department of Computing, Hong Kong Polytechnic University (PolyU), HKSAR, China (e-mail: 
xiangrong0302@gmail.com; 
jing-amelia.li@polyu.edu.hk). Jing Li is also with the PolyU Research Centre on Data Science \& Artificial Intelligence (RC-DSAI).}

\thanks{\textdagger Work done when Yixia Li and Yanlin Song were with the Department of Computing, Hong Kong Polytechnic University. 
Yixia Li is now with the Department of Statistics and Data Science, Southern University of Science and Technology (email: liyixia@me.com). Yanlin Song is now with the Department of Computing, Wuhan University (email: 2024102110023@whu.edu.cn).}



}



\maketitle             
        \begin{abstract}

Social media platforms are vital for expressing opinions and understanding public sentiment, yet many analytical tools overlook passive users who mainly consume content without engaging actively. To address this, we introduce UniPoll, an advanced framework designed to automatically generate polls from social media posts using sophisticated natural language generation (NLG) techniques. Unlike traditional methods that struggle with social media’s informal and context-sensitive nature, UniPoll leverages enriched contexts from user comments and employs multi-objective optimization to enhance poll relevance and engagement.
To tackle the inherently noisy nature of social media data, UniPoll incorporates Retrieval-Augmented Generation (RAG) and synthetic data generation, ensuring robust performance across real-world scenarios. The framework surpasses existing models, including T5, ChatGLM3 and GPT-3.5, in generating coherent and contextually appropriate question-answer pairs.
Evaluated on the Chinese \textit{WeiboPolls} dataset and the newly introduced English \textit{RedditPolls} dataset, UniPoll demonstrates superior cross-lingual and cross-platform capabilities, making it a potent tool to boost user engagement and create a more inclusive environment for interaction.


\end{abstract}

        \section{Introduction}\label{section:introduction}

\begin{figure}
    \centering
    \small
    \begin{tabular}{|l|p{6.5cm}|}
        
        
        
      \hline
        \textbf{Post}  & 据民政部统计目前中国有超 2.4 亿单身成年人，其中包括超过 7700 万独居成年人。一些独居者认为只有在与自己相处时才能得到片刻的欢愉与自在，但并不是每个人都享受一个人生活的状态。 (According to the Ministry of Civil Affairs, there are currently more than 240 million single adults in China, including more than 77 million adults living alone. Some people who live alone believe that they can only get a moment of joy and comfort when they are on their own, but not everyone enjoys living alone.)                                                        \\ \hline
        \textbf{Comments} & 我享受一个人的孤独。我现在觉得一个人看电影蛮爽的。除了吃火锅其他都做过。都做过。我觉得一个人吃餐馆。(I enjoy the solitude of being alone. How cool is it to watch a movie alone? I've done everything except eat hot pot. All have done. I feel like eating alone in a restaurant.)
        \\ 
           \hline
           \textbf{Question} & 你做过最孤独的事情是什么？(What is the loneliest thing you have ever done?) 
        \\ \hline
        \textbf{Answers}  & 一个人看电影 (Watching a movie alone); 一个人吃火锅 (Eating hot pot alone); 一个人搬家 (Moving house alone); 一个人去医院 (Going to hospital alone)                                   \\ \hline
    \end{tabular}
    \caption{
    An exemplary Weibo poll from the \textit{WeiboPolls} dataset \cite{lu_engage_2021} with the original Chinese text and its corresponding English translation within brackets. 
    The first row displays the social media post that accompanies the poll.
    The second row presents some comment samples from the post's viewers. 
    The poll extends across the 3rd and 4th rows, with the Question preceding the Answers and multiple answer choices separated by a semicolon.
    }
\label{fig:data_sample}
\vspace{-1em}
\end{figure}

\IEEEPARstart{S}{ocial} media is an essential real-time communication platform 
for expressing opinions.
It has evolved into an invaluable resource for capturing public perspectives using advanced text analytics technology, such as information retrieval and opinion mining \cite{bilal_template-based_2022}. 
This development enables people to swiftly access the latest updates and public opinions, reducing information disparities and offering unique insights into societal viewpoints.
However, many people primarily focus on consuming information rather than voicing their perspectives \cite{bi_predicting_2023}, which may result from introverted personality traits, a busy schedule, or difficulties in using text-based communication. 
It reveals the existence of the \textit{silent majority}, whose passive communication obstructs the flow of ideas and prevents others from grasping their perspective on a particular topic.
Consequently, text analytics tools may only capture a limited spectrum of public opinions, potentially overlooking the viewpoints of the silent majority and introducing biases in the interpretation of societal trends and preferences.

\IEEEpubidadjcol

To boost public engagement, numerous social media platforms have implemented polls to lower the barrier to expressing opinions. 
Users can craft the \textbf{poll} within the context of their post, and it typically consists of a question with multiple answer choices, as exampled in Fig. \ref{fig:data_sample}.
The \textbf{Question} directs readers to comprehend the post's topic and presents the primary inquiry or statement to which the post author seeks responses. 
The \textbf{Answers} comprise pre-defined choices encompassing a range of potential perspectives, enabling readers to select their viewpoints in response to the question.
Compared to traditional commenting, polls streamline the interaction process --- users can convey their opinions with a single click, bypassing the need for careful drafting and extensive typing. 
Furthermore, a recent study reveals that adding polls can helpfully enhance public engagement to a post, leading to increased comments on the Chinese Weibo platform \cite{lu_engage_2021}.

Despite the proven benefits of polls in boosting engagement \cite{lu_engage_2021}, only a small fraction of social media posts incorporate them, leaving untapped potential. Moreover, creating high-quality polls requires expertise, which not all users possess. To bridge this gap, we explore the automatic generation of polls based on social media post content. Recent advances in natural language generation (NLG), which focus on generating text conditioned on input context, make this task increasingly feasible. Modern NLG models leverage deep learning architectures that employ encoders to capture context semantics and decoders to generate coherent outputs. State-of-the-art models, such as Transformer-based architectures like GPT \cite{radford_improving_2018}, BART \cite{lewis_bart_2020}, and T5 \cite{raffel_exploring_2020}, are particularly well-suited for this task.

This work aligns with the broader research area of Question-Answer Generation (QAG), which focuses on generating question-answer pairs from a given document \cite{du_learning_2017}. QAG has been extensively studied in domains such as reading comprehension \cite{bao_unilmv2_2020} and educational applications \cite{yao_it_2022}, demonstrating strong performance in structured text generation. However, applying QAG to social media presents unique challenges due to the colloquial and contextually implicit nature of social media data, which contrasts with the explicit context-question mappings typically found in more formal language. Consequently, adapting QAG methods to social media poll generation requires addressing these implicit relationships.

For instance, the poll shown in Fig. \ref{fig:data_sample} accompanies a post discussing the prevalence of unmarried individuals living alone, yet the poll question focuses on personal experiences of loneliness. Inferring relevant answer choices, such as ``watching a movie alone," from such loosely related contexts can be difficult, as polls tend to serve as extensions of the post’s theme, prompting open-ended responses. These implicit context-question-answer relations pose significant challenges for poll generation. Prior work \cite{lu_engage_2021} demonstrated that leveraging user comments can enhance the understanding of a post’s context. Following this insight, we augment the post context with its associated comments to surface keywords indicative of user interests, such as ``alone," ``movie," and ``hotpot," thereby enriching the generated polls.

We then formulate social media poll generation following prior work \cite{lu_engage_2021}: 
the input, referred to as the \textbf{context}, comprises a \textit{post} along with its associated \textit{comments}; 
the output is a \textbf{poll}, including a \textit{question} and several \textit{answer choices}.
Our work builds upon their study with two key novelties: first, we aim to advance pre-trained NLG for social media poll generation, which has not been explored before; second, they ignored the modeling of question-answer consistency while we strive to strengthen context-question-answer relations jointly.

\textit{To the best of our knowledge, we are the first to advance pre-trained NLG models' capabilities in handling noisy social media data for poll generation.}
Our goal is to address implicit context-question-answer relations effectively.
It is thus vital to discern ``poll-worthy'' points within the contexts, incorporate diverse perspectives for crafting open-ended questions and answer choices, and manage question-answer semantic consistency.
To tackle these objectives collectively, we present a \textbf{Uni}fied \textbf{Poll} Generation framework, \textbf{UniPoll}, utilizing multi-objective optimization to balance adeptly the relation learning among contexts, questions, and answers.
UniPoll is built on T5 \cite{raffel_exploring_2020}, a state-of-the-art Seq2Seq-based pre-trained NLG model, and is trained on multiple tasks. The primary task focuses on modeling the context-poll relationship, generating a sequence with a question followed by multiple answer choices. To enhance this process, we incorporate two auxiliary tasks—question and answer generation—designed to strengthen context-question and context-answer learning. These auxiliary tasks refine the model’s ability to handle the main task, introducing diverse, targeted features that boost performance. 
\hl{Its theoretical ground comes from Pareto optimality to generate a high-quality poll without compromising question or answer quality in task balancing. Additionally, UniPoll leverages \textbf{Retrieval-Augmented Generation (RAG)}, which introduces a novel mechanism for enriching poll generation by retrieving external knowledge or related posts to supplement the input context, particularly when user comments are noisy or absent. This approach allows UniPoll to capture broader contextual signals, making it more resilient to incomplete or sparse data sources typically encountered in real-time social media analysis. Furthermore, we integrate \textbf{synthetic data generation} through large language models to simulate user interactions, offering a dynamic method for enhancing context when retrieval fails.} By adopting these strategies, UniPoll effectively manages noisy, ambiguous inputs and ensures the generation of relevant, coherent polls, thereby advancing RAG’s use in interactive social platforms. Through prompt tuning and multi-objective optimization, UniPoll fosters robust inter-task collaboration, achieving a sophisticated understanding of implicit context-question-answer relationships, surpassing standard generation techniques.

We evaluate UniPoll on the \textit{WeiboPolls} dataset \cite{lu_engage_2021}, which contains large-scale polls and their contexts from the Chinese social media platform Weibo. To further demonstrate cross-lingual generalization, we introduce a new \textit{RedditPolls} dataset, featuring English polls collected from Reddit, offering a valuable resource for future research in multilingual and cross-platform poll generation. In comparison to state-of-the-art models, including T5 and GPT-3.5-turbo, UniPoll consistently achieves superior results, as evidenced by both automatic metrics and human evaluations. Ablation studies confirm that combining the main and auxiliary tasks is key to UniPoll’s effectiveness, while sensitivity analysis highlights its robust performance across varying comment availability and data scales. Case studies underscore UniPoll’s ability to maintain strong context-question-answer consistency, and error analysis identifies areas for further refinement.

In summary, this paper presents three key contributions:

$\bullet$ We introduce UniPoll, the first framework specifically designed to enhance pre-trained NLG models for handling colloquial social media language and generating polls from noisy or incomplete contexts. Through dynamic multi-objective optimization and prompt tuning, UniPoll effectively captures implicit context-question-answer relationships.

$\bullet$ \hl{We pioneer the integration of RAG and synthetic data generation to enhance the capability of poll generation systems in managing noisy social media data and supporting practical applications. It enables cold-start poll generation without real user inputs, addressing data sparsity challenges and significantly improving performance in real-world scenarios.}

$\bullet$ Experimental results on the large-scale Chinese \textit{WeiboPolls} and \hl{our newly introduced English \textit{RedditPolls}} benchmarks demonstrate UniPoll’s state-of-the-art performance, significantly surpassing models like T5, ChatGLM3, and GPT-3.5, advancing research across languages and platforms.


The structure of this paper is outlined as follows: 
Section \ref{section:background_study} presents the related work on poll generation and multi-objective training.
Then, in Section \ref{section:methodology}, we describe the design of the UniPoll framework. The experimental setup is introduced in Section \ref{section:experiments}, with results discussed in Section \ref{section:results}. 
Finally, in Section \ref{section:conclusion}, we conclude our key findings.

        \section{Related Work}\label{section:background_study}


Our article focuses on the task of social media poll generation, building upon prior research in Question-Answer Generation (QAG) and, more specifically, Multiple Choice Question Generation (MCQG), which will be respectively discussed in Section \ref{QAG} and \ref{MCQG}.
Our method is based on multi-objective training, whose related work will be discussed in Section \ref{multi-objective}.

\subsection{Question-Answer Generation (QAG)}\label{QAG}
Poll generation aligns with \textbf{Question-Answer Generation} (QAG), a natural language processing task merging Question-Answering (QA) and Question Generation (QG). Its goal is to generate a question and corresponding answer (QA pair) from a document’s context.
According to the QAG definition, QA pairs are neither subjective nor open-ended, with answers serving as ground truth for questions \cite{du_learning_2017}, emphasizing explicit QA-context connections.
QAG is widely used in education \cite{yao_looking_2022} for automated assignment or exam questions.It is also a valuable approach for data augmentation by self-labeling large-scale data for training reading comprehension \cite{bao_unilmv2_2020,du_harvesting_2018}.

Early research focused on collaboratively training QA and QG models, leveraging their interdependence for mutual improvement. Studies have shown that posing questions while reading enhances comprehension, yielding better answers \cite{singer_active_1982}. Similarly, generating questions while answering helps models discern QA connections.
Building on these findings, researchers explored techniques for collaborative training of QA and QG models. 
Tang \textit{et al.} \cite{tang_question_2017} employed GRU and Bi-GRU models for joint training of discriminative QA and generative QG. They framed QA and QG as dual tasks, positing a probabilistic relationship.
By integrating probabilistic duality constraints into loss functions, they enhanced QA and QG performance on benchmarks like \emph{MARCO} \cite{nguyen_ms_2016}, \emph{SQuAD} \cite{rajpurkar_squad_2016}, and \emph{WikiQA} \cite{yang_wikiqa_2015}. These results highlight the potential of collaborative training for QAG models.

An alternative approach involves Generative Domain-Adaptive Nets (GDAN \cite{yang_semi-supervised_2017}) and Generative Adversarial Networks (GAN, \cite{goodfellow_generative_2014, wang_irgan_2017}). Tang \textit{et al.} \cite{tang_learning_2018} used these techniques to enhance QG by introducing QA-specific signals into the loss function. Inspired by GANs, they utilized QG-generated samples for QA training, further boosting QAG performance.

Nonetheless, utilizing two separate models (QA and QG) can be inefficient regarding time and computational resources. 
It may also pose challenges in coordinating their contributions to QAG tasks. 
To address the concern, Wang \textit{et al.} \cite{wang_joint_2017} suggested an integrated seq2seq framework for both QA and QG, incorporating a condition encoder to toggle between the two tasks. 
Recent advancements in pre-training have further enhanced the popularity of integrated QAG models.  
Shakeri \textit{et al.} \cite{shakeri_end--end_2020} employed a Seq2Seq framework based on the pre-trained BART transformers \cite{lewis_bart_2020}.
They proposed a two-step, two-pass method: the system first generates a question, which is later combined with the context for a second-pass answer generation.
A more recent work Yao \textit{et al.} \cite{yao_it_2022} adopted heuristic searching to extract candidate answers and passed them through a BART-based QG module to generate questions. 

Our work adheres to the state-of-the-art practice of adopting a unified QAG model, with a pre-trained language model as the backbone.
However, previous work mainly focuses on the QAG for formal documents, such as Wikipedia, which exhibits clearer relations among the context, question, and answers. 
However, social media language usually exhibits a much more implicit context-question-answer relation, which may compromise existing NLG models' performance.
Here we present a novel solution with task prompts to distinguish relation learning among contexts, questions, and answers, and multi-objective training to strengthen the capturing of their joint connections.
In addition, the typical QAG often pertains to text-based answers rather than answer choices, differentiating it somewhat from a poll.
Our task aligns more closely with the specialized domain of Multiple Choice Question Generation (MCQG), which we will discuss in the following subsection.

\subsection{Multiple Choice Question Generation (MCQG)}\label{MCQG}
As introduced above, a poll consists of a question and several answer choices, similar to a \textbf{Multi-Choice Question} (MCQ).
This characteristic makes our task specifically related to Multi-Choice Question Generation (MCQG).
Most prior MCQG studies explored the MCQ for
assessments, which
presents a question with a correct answer (also known as the key) and several incorrect answers (also known as distractors). 
It is valuable in education due to MCQs' consistent scoring and easy evaluation
\cite{brown_automatic_2005, sumita_measuring_2005, chen_fast_2006, mostow_generating_2009}. 
Besides, some work also applied MCQG to evaluate readers' news comprehension \cite{lelkes_quiz-style_2021}. 

As for methodology, many early MCQG systems employ heuristic or rule-based generation \cite{ch_automatic_2020}. 
Some also exploited structured knowledge graphs to encode the knowledge for assessments \cite{alsubait_generating_2014, seyler_knowledge_2017}. 
In addition, others specifically explored crucial sub-problems in key selection and question formation.

In key selection, a keyword or sentence is selected to serve as the answer for the subsequent question generation. 
Traditional approaches \cite{karamanis_generating_2006, faizan_automatic_2018} to this task, such as the Tf-IDF algorithm, rely on statistical and lexical information. 
These methods often struggle to identify keys with referential relationships and can only produce simple, straightforward words. 
To address these issues, researchers have incorporated semantic information \cite{afzal_automatic_2014, fattoh_automatic_2014} and employed machine learning techniques \cite{lelkes_quiz-style_2021,welbl_crowdsourcing_2017} to extract longer phrases or sentences as answers, substantially improving MCQG performance.

In question formation, a question is generated based on the selected key. 
Traditional methods for this task rely on rules and syntactic dependencies \cite{afzal_automatic_2014, majumder_system_2015}, which often cannot well model the context, compromise the performance on out-of-domain data, and result in lack-of-diversity output.
Viewing these limitations, recent approaches adopted neural network-based sequence-to-sequence models, some even with the pre-trained encoder-decoder \cite{vaswani_attention_2017-1,bahdanau_neural_2015}, such as T5  \cite{raffel_exploring_2020} and PEGASUS \cite{zhang_pegasus_2020}. 
In particular, Hadifar \textit{et al.} \cite{hadifar_eduqg_2022} fine-tuned T5 on the \emph{SQuAD} \cite{rajpurkar_squad_2016} and continued to fine-tune it on  \emph{EduQG}\cite{hadifar_eduqg_2022}, exhibiting improved quality in the generated questions.

In the MCQG pipeline, previous work usually searched for a target answer and generated questions based on the context to find potential question subjects.
Although key selection and question formation are closely-related tasks, they were priorly usually tackled separately in MCQG systems, resulting in inferiority in learning their interactions. 
As the simultaneous training of key and question showed better performance than the pipeline alternative \cite{tang_question_2017}, the Seq2Seq-based T5 and PEGASUS were successfully applied to MCQG systems in the latest study \cite{lelkes_quiz-style_2021, raina_multiple-choice_2022, vachev_leaf_2022} for sequential generations of key and question. 
Built upon that, we adopt T5 as our backbone and generate poll question and answer choices sequentially. 
However, previous MCQG approaches do not explicitly model the relationship between the question and the answers during training, which may limit their performance in tackling noisy social media polls with implicit context-question-answer relations. To address this problem, we incorporate two auxiliary tasks to better coordinate the answer and question generation.

Despite being in the MCQ format, a social media poll fundamentally differs from an assessment-oriented MCQ --- the former is open-ended, used for opinion collection, and adopts a colloquial style, while the latter is to test an individual's knowledge, includes a specific correct answer, and is typically written formally.
Lu \textit{et al.} \cite{lu_engage_2021}, being the only prior work in \textbf{social media poll generation}, contributed the first Chinese dataset for this task (\textit{WeiboPolls} for short). 
They gathered polls from Chinese social media, Sina Weibo\footnote{\href{https://www.weibo.com}{https://www.weibo.com}}.
Their method involved two decoders based on Bi-GRU \cite{cho_learning_2014}, one for question generation and the other for answer choices.
They also employed a neural topic model \cite{miao_discovering_2017, zeng_topic_2018} to effectively model the comments to enrich the contexts for input understanding.
However, they did not adopt a pre-trained decoder to harness the NLG advances in pre-trained language models. 
In addition, they processed questions and answers using separate decoders without comprehensively exploring question-answer interactions. 
In contrast, our model is grounded on a pre-trained decoder and leverages multi-objective training to examine the context-question-answer relations thoroughly. 
\hl{Recently, Cheng \textit{et al.} }\cite{cheng2024diffuspoll}\hl{ introduced DiffusPoll, a conditional text diffusion model for poll generation. Similar to UniPoll, it employs task-specific strategies to ensure coherence between questions and options. However, DiffusPoll builds on extensive resources, UniPoll is more lightweight yet effective. Also, DiffusPoll enhances data with back-translation and attribute tag extraction; our approach focuses more on modeling implicit relations.}

\subsection{Multi-Objective Optimization}\label{multi-objective}

Our method is based on established research in multi-objective optimization to coordinate the objectives to generate questions and answers in polls.
In previous work, this approach is generally employed to optimize the problem involving more than one objective function, resulting in the popularity to apply in multi-task learning \cite{sener_multi-task_2018}.
Here, different tasks can reach Pareto stationary points by introducing the Multiple Gradient Descent Algorithm (MGDA) from multi-objective optimization. 
Later work further refined this idea by optimizing task weights and computation efficiency \cite{lin_pareto_2019, ma_efficient_2020, momma_multi-objective_2022}.

Building on the foundation of multi-objective optimization, our training design engages multiple tasks in a unified framework, enabling joint handling of different objectives. This approach is inspired by its demonstrated success in dialogue generation, where it promotes knowledge transfer between tasks by sharing parameters and feature representations, enhancing a model’s generalization capabilities over single-task learning \cite{gao_unigdd_2022}. For instance, the T5 model explores data from various tasks during pre-training, significantly improving out-of-distribution capabilities, as seen in FLAN-T5, which scales the number of fine-tuning tasks to 1,800 \cite{raffel_exploring_2020, chung_scaling_2022}.

In contrast to traditional multitask learning frameworks \cite{lu_engage_2021} that often require specific encoders for each task, limiting flexibility and applicability, our innovative use of prompt tuning enables a more versatile approach to task differentiation. This flexibility allows seamless integration of our method into existing encoder-decoder architectures like FLAN-T5 or even decoder-only models such as GPT-based ones. By employing prompt-based differentiation, our framework adapts naturally to varying language models, enhancing usability and adaptability across architectural configurations.

\lyx{However, despite these advancements, prior work has rarely explored the effects of multi-objective optimization in managing the implicit context-question-answer relations specific to social media polls. Addressing this gap, our model incorporates both main and auxiliary tasks to improve understanding and generation of context-poll relationships, further facilitated by the distinct task prompts that help delineate and connect various objectives within the model.}
        \section{UniPoll Model for Poll Generation} \label{section:methodology}

In this section, we first formulate the social media poll generation task, followed by three subsections to elaborate on how we design our UniPoll framework with the overview in Section \ref{ssec:framework}, main and auxiliary task design in Section \ref{ssec:task-design}, and multi-objective optimization in Section \ref{ssec:multi-objective-training}.

\begin{table}
\centering
\caption{Math Notations.\label{table:notions}}
\begin{tabular}{ll}
\hline
Notation & Description                                                \\ \hline
$TP$       & the task prompt                                            \\
$P$        & the source post                                            \\
$C_n$      & the $n$ comments related to the source post                \\
$X$        & the input word embedding                                   \\
$M$        & the encoded memory bank from encoder                       \\
$Q$        & the poll question                                          \\
$A_m$      & the poll answers, with $m$ answers                         \\
$N$        & the number of training samples                             \\
$\gamma_Q$ & the loss weight of auxiliary task QG                       \\
$\gamma_A$ & the loss weight of auxiliary task AG                       \\ \hline
\end{tabular}
\end{table}

Following the task introduction in Section \ref{section:introduction}, we further detail the social media poll generation task formulation.
The given context (input) is a social media \textbf{Post} $P$ that presents a topic or a statement as the background and its comments $C_n$ as supplementary contexts with some user views, where $n$ indicates the comment num
Our output is a poll with a \textbf{Question} $Q$, to seek other users' opinions, and $m$ \textbf{Answers} $A_m$, a sequence of choices to provide varying potential perspectives for users to select as their standpoint.
Our task thus involves both question generation (QG) and answer generation (AG). The notions used in this section are briefed in Table \ref{table:notions}.

\subsection{UniPoll Framework Overview}\label{ssec:framework}

As discussed above, UniPoll adopts an encoder-decoder architecture based on the pre-trained Seq2Seq model T5. This architecture leverages the model's capabilities in understanding and generating natural language, which is crucial for handling informal and diverse expressions in social media.
%
%
The key challenge lies in the implicit relationships between context, questions, and answers on social media platforms. To address the challenge, UniPoll employs a multi-objective optimization strategy that integrates multiple tasks while balancing their respective goals throughout the learning process. 

\hl{Despite the simple design, UniPoll's theoretical ground comes from Pareto-optimal task balancing, facilitating a trade-off between conflicting tasks} \cite{huang-etal-2023-towards}. In multi-task learning, tasks often exhibit competing gradients, and optimizing for one task can hinder the performance of others. By leveraging the Pareto optimality criterion, the model identifies solutions where improving one task does not significantly degrade others. Gradient-based multi-objective optimization methods, such as decomposing tasks into subproblems with different preferences, guide the search through more well-behaved, representative solutions. This structured search not only leads to improved task coordination but also helps in stabilizing the optimization process \cite{lin2019paretomultitasklearning}. By mitigating conflicts between task gradients, the approach indirectly contributes to smoother convergence and a more stable loss function, enabling the model to learn more effectively. As a result, UniPoll is able to generate high-quality polls without compromising the quality of the questions or answers. By focusing on the primary task of poll generation alongside two auxiliary tasks—refining the formulation of both questions and answers—UniPoll achieves a well-coordinated balance between these objectives, enhancing both convergence stability and output coherence.

Furthermore, UniPoll employs a task-aware gradient interaction mechanism, enabling gradients from different tasks to synergize and improve the model’s overall performance. This synergy is achieved through carefully designed task-specific prompts, guiding the model to consider both question generation (QG) and answer generation (AG) in a complementary manner. By facilitating the sharing of information across tasks, UniPoll enhances the consistency and coherence of its outputs. This prompt-driven, gradient-based multi-task learning strategy balances competing objectives, ensuring robust and well-rounded performance across all tasks.

\hl{To address the noisiness of social media data, UniPoll leverages two approaches: Retrieval-Augmented Generation (RAG)} \cite{gao2024retrievalaugmented}\hl{ and synthetic data generation using large language models. RAG enables the model to retrieve relevant information from external sources, enriching the context when user comments are unavailable or of low quality. On the other hand, synthetic data generation employs large language models to create artificial comments, providing additional context in scenarios where real comments are missing or insufficient. These techniques allow UniPoll to handle both noisy and incomplete data, enhancing the quality of the generated polls.}


\begin{figure}
    \centering
    \includegraphics[width=7.5cm]{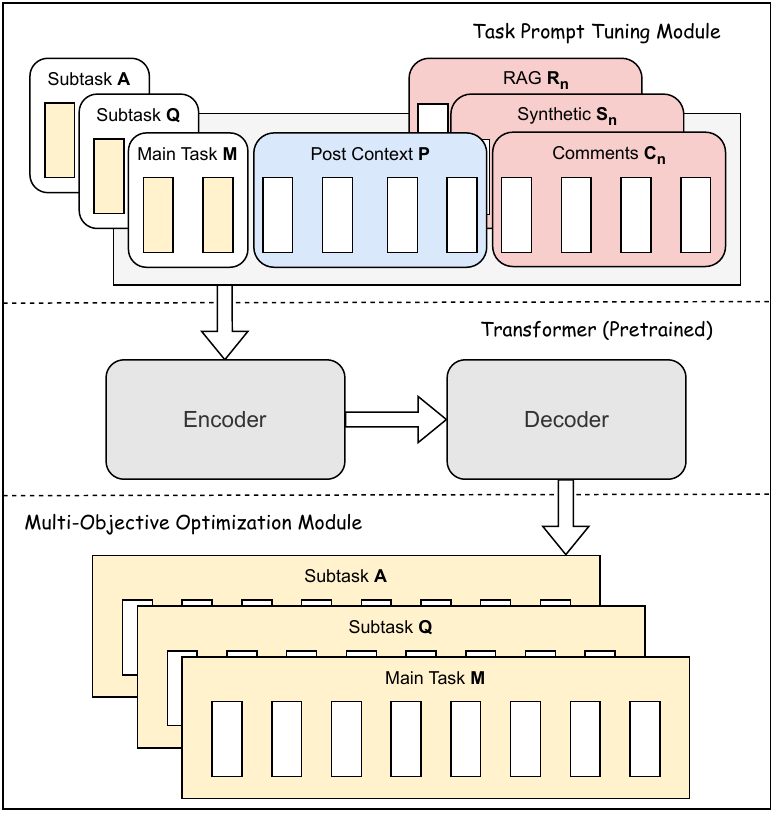}
    \vspace{-0.25cm}
    \caption{The overview of UniPoll, the unified poll generation framework. The middle displays the backbone, a pre-trained transformer (T5) in an encoder-decoder architecture. The top shows the task prompting design to help the model differentiate tasks and the bottom part is our multi-objective training to coordinate the relation learning for contexts, questions, and answers.}    
    \label{fig:poll_question_framework}
\end{figure}

The overview and operational intricacies of UniPoll are visually depicted in Fig. \ref{fig:poll_question_framework}\hl{, which illustrates how UniPoll processes input data to generate well-structured and contextually meaningful polls.}
\lyx{Through this multi-faceted approach, UniPoll aims to capture the \textit{mutual dependence of QG and AG}. This interdependence is critical, as the nature of questions can influence the type of answers provided and vice versa. For example, a question about a location (``Where") requires answers that are places, whereas a question about a person (``Who") necessitates names as responses. By tailoring the model to consider these aspects, UniPoll can generate more coherent and contextually relevant content, thereby significantly improving the quality of interaction on social media platforms.}

\subsection{Main and Auxiliary Task Design\label{ssec:task-design}}

We train UniPoll with multiple tasks to better coordinate QG and AG with the implicit context-question-answer relations.

\paragraph{Main Task} 
The main task handles the main quest to generate a poll directly from the context.
Its input consists of three components: task prompt $TP$, post context $P$, and comments $C_n$, combined by the separator token. 
Here we introduce $TP$ to help the model discriminate between specific tasks. 
For the main task targets, we employ task prompt $TP_M$ \textit{``generate \textless question\textgreater \  then \textless answers\textgreater "}, where \textit{``\textless question\textgreater "} and \textit{``\textless answers\textgreater "} are two additional special tokens indicating the question and answers. 

The input triplet $<TP, P, C_n>$ is formed as a sequence of word embeddings $X$ and fed into the encoder to generate the memory bank $M$.  The process of generating the memory bank can be represented by the following formula:

\begin{equation}
    M = Encoder(X)
    \label{eq:encoder}
\end{equation}

\noindent where $Encoder$ is the encoder function responsible for encoding the input text into a compact representation that captures relevant context information.
The output $M$ is the memory bank, which is a matrix where each column represents the hidden state of the encoder for a particular token.

Guided by $TP_M$, the main task outputs a sequence of the poll question, concatenated by its answer choices, with each distinct target preceded by a special token. 
The transformer decoder consists of a series of decoder layers to digest the general semantics in the input context and exploit it to generate the output poll.
To that end, the decoder employs a multi-headed attention mechanism to attend to both the input sequence and the memory bank $M$. It enables the decoder to incorporate a salient context part when predicting each output word. 
For inference, the transformer decoder generates a question $Q$ and answers $A_m$ by processing a sequence of input tokens, one at a time, and conditioned on the previously generated tokens to predict the next token in the sequence.  
The generation of poll question $Q$ can be formulated by the following equation:

\begin{equation}
    Pr(Q\ |\ TP_M, P, C_n) = \prod_{j=1}^{|\mathbf{q}|} Pr(q_j|\mathbf{q}_{<j}, M)
    \label{eq:main_q}
\end{equation}

Here $q_j$ denotes the $j$-th word in $Q$ and $\mathbf{q}_{<j}$ refers to $Q$'s predicted word sequence from slot $1$ to $j-1$. 
Upon reaching the special separate token, the decoder automatically switches to answer generation. The answer-generation process incorporates information from both the memory bank $M$ and the generated question $Q$, enabling it to function in a question-aware manner, as shown below:

\begin{equation}
    Pr(A_m \ |\ TP_M, P, C_n) = \prod_{j=1}^{|\mathbf{a}|} Pr(a_j|\mathbf{a}_{<j}, Q, M)
    \label{eq:main_a}
\end{equation}

For a better illustration, Fig. \ref{fig:formatted_example} shows how we format the input and output Fig. \ref{fig:data_sample} example for the main task.
It maps the inputs $<TP_M, P, C_n>$  to the outputs $<Q, A_m>$ with the encoder-decoder model in a Seq2Seq manner. During training, the main task is involved as one of the tasks, and during inference, only the main task will be used for prediction.

\begin{figure}
\begin{tcolorbox}[colback=gray!5!white, colframe=black!75!black]
    \textbf{Input}: \textcolor{orange}{\textit{generate \textless question\textgreater \ then \textless answers\textgreater }:} [sep] \textcolor{blue}{据民政部统计目前中国有超 2.4 亿单身成年人...} [sep] \textcolor{red}{我享受一个人的孤独。我现在觉得一个人看电影蛮爽的。都做过。我觉得一个人吃餐馆。}
    
    \textbf{Output}: \textcolor{orange}{\textit{\textless question\textgreater }} 你做过最孤独的事情是什么？ [sep] \textcolor{orange}{\textit{\textless answers\textgreater }} 一个人看电影 [sep] 一个人吃火锅 [sep] 一个人搬家 [sep] 一个人去医院
\end{tcolorbox}
\caption{The formatted input and output of the Fig. \ref{fig:data_sample} example for the main task. To distinguish different parts, the \textcolor{orange}{prompt prefix} and \textcolor{orange}{special tokens} are in orange color; the \textcolor{blue}{post context} and \textcolor{red}{comments} are in blue and red, respectively.}
\label{fig:formatted_example}
\end{figure}

In addition to the intuitive Seq2Seq model, we retain the original cross-entropy loss function. Instead of formulating a distinct loss function, the model calculates the joint loss of both QG and AG simultaneously to choose an optimal equilibrium point, which will be detailed in Section \ref{ssec:multi-objective-training}

\paragraph{Auxiliary Tasks} 
Although the main task allows poll generation, the model may over-learn one target (QG or AG) while under-fitting the other if only considering the joint loss being computed in the hidden layer. 
To mitigate this issue, UniPoll engages two side quests to explicitly learn QG and AG as auxiliary tasks to optimize the fitting degrees for QA and QG.
They enable target-specific learning to align the related context specifically with the question or answers, facilitating the main task of reinforcing the connections between context-question and context-answer pairs. 
It allows for a more balanced training, as the model can explicitly discern the training degrees locally for QG and AG, meanwhile reinforcing the global learning for context-question-answer relations.


The input of auxiliary tasks is the triple $<TP, P, C_n>$,  while the output is differentiated by task prompts. 
We designed two task prompts: $TP_Q$ \textit{``generate \textless question\textgreater "} and $TP_A$ \textit{``generate \textless answers\textgreater "}, which correspond to QG and AG, respectively. 
The prompt templates are crafted to relate auxiliary tasks to the main task. 
By jointly tackling main and auxiliary tasks, UniPoll can not only distinguish the fitting magnitude of varying targets but also learn the connections between QA pairs simultaneously.
The output generation of auxiliary tasks is formalized as follows: 

\begin{equation}
    Pr(Q\ |\ TP_Q, P, C_n) = \prod_{j=1}^{|\mathbf{q}|} Pr(q_j|\mathbf{q}_{<j}, M)
    \label{eq:Q_q}
\end{equation}

\begin{equation}
    Pr(A_m \ |\ TP_A, P, C_n) = \prod_{j=1}^{|\mathbf{a}|} Pr(a_j|\mathbf{a}_{<j}, M)
    \label{eq:A_a}
\end{equation}

\subsection{Multi-Objective Optimization}\label{ssec:multi-objective-training} 

With the previously constructed task prefix $TP$, the main task and two auxiliary tasks can be trained simultaneously under the encoder-decoder model, while only the main task is used to generate questions and answers during inference.

The main task employs the following training loss:

\begin{equation}
    \mathcal{L}_{Main} = - \sum_{i=1}^N log(Pr(Q^i, A^i_m \ |\ TP_M, P^i,C_n^i))
    \label{eq:main_loss}
\end{equation}

\noindent $N$ is the number of training samples; $P^i$, $C_n^i$ are the source post and its comments of the $i$-th training sample, and $Q^i$, $A^i_m$ are the target poll question and poll answers. 
For the auxiliary tasks, the QG and AG exhibit the following losses:

\begin{equation}
    \mathcal{L}_{Q} = - \sum_{i=1}^N log(Pr(Q^i \ |\ TP_Q, P^i,C_n^i))
    \label{eq:Q_loss}
\end{equation}

\begin{equation}
    \mathcal{L}_{A} = - \sum_{i=1}^N log(Pr(A^i_m \ |\ TP_A, P^i,C_n^i))
    \label{eq:A_loss}
\end{equation}

To enable multi-objective optimization for coordinating context-poll, context-question, and context-answer relation learning, the training loss of the entire model is defined as:

\begin{equation}
    \mathcal{L} = \mathcal{L}_{Main} + \gamma_Q \cdot \mathcal{L}_{Q} + \gamma_A \cdot \mathcal{L}_{A} 
    \label{eq:overall_loss}
\end{equation}


\noindent where weights $\gamma_Q$ and $\gamma_A$ respectively trade-off main and auxiliary tasks for QG and AG.
Eq. \ref{eq:overall_loss} is optimized to coordinate multiple tasks' objectives with Algorithm \ref{algorithm}'s workflow.


\begin{algorithm}
\small
\caption{Multi-Objective Optimization for UniPoll}\label{algorithm}
\begin{algorithmic}[1]
\renewcommand{\algorithmicrequire}{\textbf{Input:}}
\renewcommand{\algorithmicensure}{\textbf{Output:}}
\REQUIRE A set of training samples $\{D\}$, subtask options (Main, QG, AG), loss weights $\gamma_Q, \gamma_A$, learning rate $\eta$, and maximum iterations $maxIter$
\ENSURE Optimized model parameters $\theta$
    \STATE \textbf{Initialize:} Model parameters $\theta$, learning rate $\eta$
    \STATE Generate training data $\{D'\}$ with subtask options
    \FOR {epoch from $1$ to $maxIter$}
        \STATE Partition $\{D'\}$ into batches
        \FOR {each batch $b \in \{D'\}$}
            \STATE Extract source post $P$, comments $C_n$, and ground truth $Q$, $A_m$ from $b$
            \STATE Embed the input triplet $\mathbf{X} = <TP, P, C_n>$ and pass through the encoder to generate memory bank $M$:
            \[
                M = \text{Encoder}(\mathbf{X})
            \]
            \vspace{-8pt}
            \STATE Decode to generate the poll question $Q$ and answers $A_m$:
            \[
                Pr(Q \ | \ TP_M, P, C_n) = \prod_{j=1}^{|\mathbf{q}|} Pr(q_j \ | \ \mathbf{q}_{<j}, M)
            \]
            \[
                Pr(A_m \ | \ TP_M, P, C_n) = \prod_{j=1}^{|\mathbf{a}|} Pr(a_j \ | \ \mathbf{a}_{<j}, Q, M)
            \]
            \STATE Compute the joint loss for the main and auxiliary tasks:
            \[
                \mathcal{L} = \mathcal{L}_{Main} + \gamma_Q \cdot \mathcal{L}_{Q} + \gamma_A \cdot \mathcal{L}_{A}
            \]
            \vspace{-8pt}
            \STATE Update model parameters:
            \[
                \theta = \theta - \eta \cdot \nabla_\theta \mathcal{L}
            \]
        \ENDFOR
    \ENDFOR
\end{algorithmic}
\end{algorithm}

\subsection{Automated Context Enrichment}
\begin{figure}
    \centering
    \includegraphics[width=7.5cm]{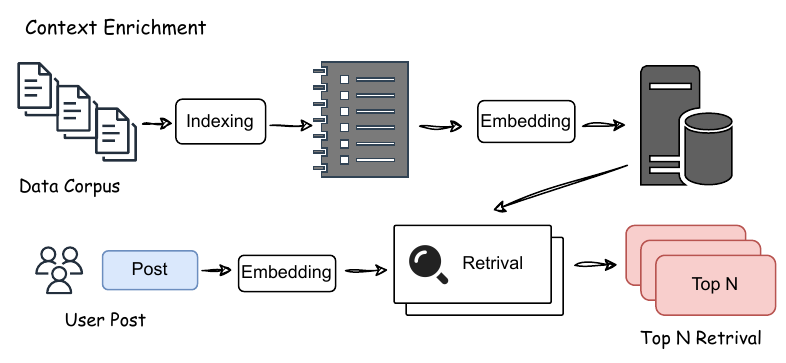}
    \vspace{-0.25cm}
    \caption{\hl{Retrieval-Augmented Generation module for context enrichment}}    
    \label{fig:poll_question_framework_rag}
\end{figure}

\hl{To further address scenarios where comments are of low quality or unavailable, UniPoll incorporates Retrieval-Augmented Generation (RAG)} \cite{gao2024retrievalaugmented} techniques to enhance context enrichment. Here, we discuss how we employ the RAG technique in detail. By augmenting the post context with additional relevant information retrieved from a large corpus, RAG allows for effective poll generation without relying on real user comments. The process is shown in Fig. \ref{fig:poll_question_framework_rag}

First, the model builds embeddings for the post $P$ using a pre-trained embedding model:

\begin{equation}
    \mathbf{E}_P = \text{Embedding}(P)
    \label{eq:post_embedding}
\end{equation}

\noindent where $\mathbf{E}_P$ represents the embedding vector of the post $P$.

Next, we retrieve relevant external contexts $R_n$ from a large corpus by calculating the similarity between the post embedding $\mathbf{E}_P$ and the embeddings of documents in the corpus formulated as follows:

\begin{equation}
    R_n = \arg\max_{D \in \text{Corpus}} \text{Sim}(\mathbf{E}_P, \mathbf{E}_D)
    \label{eq:retrieval}
\end{equation}

Here, $\text{Sim}(\mathbf{E}_P, \mathbf{E}_D)$ represents a similarity measure (e.g., cosine similarity) between the embedding of the post $P$ and the documents $D$ in the corpus. The top-$n$ documents $R_n$ are retrieved based on their similarity scores.

These retrieved contexts $R_n$ are then used in place of the comments $C_n$ for poll generation. The input triplet for the main task becomes $<TP, P, R_n>$, and the corresponding poll generation probabilities for the question and answers are formulated as:

\begin{equation}
    Pr(Q\ |\ TP_Q, P, R_n) = \prod_{j=1}^{|\mathbf{q}|} Pr(q_j|\mathbf{q}_{<j}, M)
    \label{eq:Q_rag}
\end{equation}

\begin{equation}
    Pr(A_m \ |\ TP_A, P, R_n) = \prod_{j=1}^{|\mathbf{a}|} Pr(a_j|\mathbf{a}_{<j}, M)
    \label{eq:A_rag}
\end{equation}

By incorporating the retrieved contexts $R_n$, the model effectively generates poll questions and answers without the need for user comments, leveraging external data to enrich the input context and ensure high-quality outputs.
        \section{Experimental Setup} \label{section:experiments}

This section discusses how we set up our experimental studies as follows.
First, Section \ref{subsection:dataset} discusses datasets and evaluation metrics.
Then, we describe multiple variations of the comparison methods and the configuration of our ablation models in Section \ref{subsection:comparing}.
At last, we delve into the specific implementation details of UniPoll in Section \ref{subsection:implementation}.

\subsection{Evaluation Dataset and Metrics}\label{subsection:dataset}

Here we discuss the dataset for training and testing, followed by the evaluation metrics to measure the output quality.

\paragraph{Dataset} 
We evaluate the UniPoll model using two datasets: the Chinese \textit{WeiboPolls} dataset \cite{lu_engage_2021} and the English \textit{RedditPolls} dataset. The \textit{WeiboPolls} dataset contains 20,252 poll question pairs sourced from the Chinese social media platform Weibo. Each entry includes the context of a post, its comments, and a user-generated poll with a question and a set of answers. \hl{We use the original dataset}\footnote{\href{https://github.com/polyusmart/Poll-Question-Generation/tree/main/data/Weibo}{https://github.com/polyusmart/Poll-Question-Generation/tree/main/data/Weibo}} \hl{as published by previous work}\cite{lu_engage_2021}\hl{, without any additional filtering, consistent with prior work.}
\hl{To further evaluate the generalizability of our method,} we collected English poll data from Reddit’s \textit{r/polls}\footnote{\href{https://www.reddit.com/r/polls}{https://www.reddit.com/r/polls}} using the Pushshift Reddit API\footnote{\href{https://github.com/pushshift/api}{https://github.com/pushshift/api}}. \hl{Due to API limitations, we only gathered data up until November 28, 2021, resulting in a total of 18,000 samples.}


\paragraph{Evaluation Metrics}
To assess the effectiveness of the poll generation models, we adopt ROUGE and BLEU scores \cite{lin_rouge_2004, papineni_bleu_2002, post_call_2018}, as suggested by previous work  \cite{lu_engage_2021}.   
Specifically, we follow them to employ the pythonrouge package\footnote{\href{https://github.com/tagucci/pythonrouge}{https://github.com/tagucci/pythonrouge}} to calculate ROUGE-1, ROUGE-L, and NLTK\footnote{\href{https://pypi.org/project/nltk/}{https://pypi.org/project/nltk/}} to calculate BLEU-1 and BLEU-3 scores. 
We evaluate the model's performance on three tasks: poll generation, question generation, and answers generation. The poll generation results are obtained by averaging the results of the last two tasks over a single run.

\subsection{Comparing Methods} \label{subsection:comparing}

We consider the following baselines and the existing state-of-the-art models in the experimental comparisons.

We first consider the Seq2Seq (S2S) baselines without a pre-trained backbone.
BASE \cite{sutskever_sequence_2014} is the basic RNN-based \underline{S2S} model that trains from scratch.
We also report the results of two \lyx{BERT family-based} extensions that incorporate additional features. 
\underline{COPY} \cite{meng_deep_2017} utilizes a copy mechanism to enable extractions from the context for poll generation.
\underline{TOPIC} \cite{wang_topic-aware_2019}, based on COPY, additionally incorporates topic modeling to examine word statistics in comments to understand the noisy context better.

Besides, we adopt the poll generation model released in previous work \cite{lu_engage_2021}, which is based on TOPIC yet employs a dual decoder (henceforth \underline{DUAL DEC}), one for question and one for answer generation.
It is the previous state-of-the-art model in the WeiboPoll benchmark. \lyx{DUAL DEC did not adopt the pretrain model since no performance gain with BERT, according to } \cite{lu_engage_2021}. 
To examine the advanced pre-trained NLG model in our task, we adopt a \underline{T5} baseline and follow DUAL DEC to
train two single-objective models for question and answer generation, respectively. \lyx{Specifically, we fine-tuned the T5 model separately for the question generation and answers generation tasks. After fine-tuning, we evaluated the performance of each task independently and calculated the overall score for poll generation by averaging the scores from both tasks.}
Additionally, we explore both open-source and proprietary state-of-the-art large language models. Specifically, we fine-tune the ChatGLM3-6B model\footnote{\href{https://huggingface.co/THUDM/chatglm3-6b}{https://huggingface.co/THUDM/chatglm3-6b}} \cite{du2022glm} following the same procedure used for the T5 baseline, leveraging LoRA (Low-Rank Adaptation) \cite{hu2022lora} for efficient fine-tuning. LoRA is set at a rank of 64, resulting in 15,597,568 trainable parameters out of a total of 6,259,181,568 parameters for ChatGLM3-6B. In comparison, the T5 model contains a total of 220,000,000 parameters.
\hl{For further benchmarking, we employ GPT-3.5-turbo}\footnote{\href{https://platform.openai.com/docs/models/gpt-3-5-turbo}{https://platform.openai.com/docs/models/gpt-3-5-turbo}} \hl{in a one-shot learning setup for comparison.}



In the comparison, UniPoll refers to our full model with multi-objective optimization over the main and auxiliary tasks. 
To further examine each task's relative contributions, we design UniPoll's ablations as follows. 
The \underline{w.o.A} is the ablation without answer generation auxiliary task; likewise, \underline{w.o.Q} is the ablation without question generation auxiliary task;
\underline{w.o.Q,A} only employs the main task for the poll generation. 

\subsection{Implementation Details} \label{subsection:implementation}
We used the open-source Unbuntu 20.04.5 operating system. 
All experiments were conducted with PyTorch 1.12 \cite{paszke_pytorch_2019} and Python 3.8. 
For hardware, we used two RTX 2080ti graphics cards with 11G memory each for training and inference.

Due to the GPU memory constraints, we employed the base version of the T5 pre-trained model\footnote{\href{https://huggingface.co/google-t5/t5-base}{https://huggingface.co/google-t5/t5-base}} \cite{raffel_exploring_2020} as our backbone for \textit{RedditPolls} and T5-pegasus pre-trained checkpoint \cite{su_t5_2021} for \textit{WeiboPolls}\footnote{\href{https://huggingface.co/imxly/t5-pegasus}{https://huggingface.co/imxly/t5-pegasus}}.
T5-pegasus used mT5 \cite{xue_mt5_2021} as the base architecture and customized the tokenizer for Chinese, with a pre-training task similar to PEGASUS \cite{zhang_pegasus_2020}.


For training, we adopted the AdamW \cite{loshchilov_decoupled_2019} optimizer with an initial learning rate of 3e-5 and a linear learning rate decay strategy. 
The number of training epochs is 20 for the single-task models and 10 for the multi-task models. 
The weights to trade off multi-task learning losses are set to $\gamma_Q=\gamma_A=1$ (Eq. \ref{eq:overall_loss}).
ROUGE-1 was observed in validation, and for multi-objective models (w.o.A, w.o.Q, and UniPoll), we averaged the ROUGE-1 for the generated questions and answers.

We employed beam search with a beam size of 1 for text decoding during inference. 
The maximum input length for the model is 1,024 tokens, with any input exceeding this length being truncated, and the maximum output length is 128 tokens. 
All experiments ran five times (with random seeds 40, 41, 42, 43, 44), and the mean will be reported in the results.

        \section{Experimental Results}\label{section:results}


To evaluate the performance of UniPoll, we conducted extensive experiments whose results are discussed as follows.
First, we compare UniPoll with existing baselines and state-of-the-art models for poll generation
in Section \ref{subsection:overall_performance}
, \hl{followed by the context enrichment techniques in Section} \ref{subsection:context_aug}
Subsequently, Section \ref{subsection:ablations} will provide the ablation studies to examine the effects of auxiliary tasks and multi-objective optimization for strengthening context-question-answer interrelationships.
After that, we quantify the effects of varying comment scales and training data scales to examine UniPoll's performance with varying input (in Section \ref{subsection:quantitative-analysis}).
Finally, to provide more insights on the potential and limitations, we qualitatively discuss UniPoll through a case study (in Section \ref{subsection:case_study}) and an error analysis (in Section \ref{subsection:error_analysis}).

\begin{table}
\centering
\caption{Comparison results on \textit{WeiboPolls} for poll generation. All baselines are fine-tuned, except GPT-3.5-turbo prompted in a one-shot manner. UniPoll significantly outperforms all other models (paired t-test; p-value \textless 0.05).}
\label{tab:main_results_PG}
\resizebox{1\columnwidth}{!}
{
\begin{tabular}{|l|rrrr|}
\hline
\multirow{2}{*}{\textbf{Model}} & \multicolumn{4}{c|}{\textbf{Poll Generation}}                                                                                                            \\
                                & \multicolumn{1}{c}{\textbf{ROUGE-1}} & \multicolumn{1}{c}{\textbf{ROUGE-L}} & \multicolumn{1}{c}{\textbf{BLEU-1}} & \multicolumn{1}{c|}{\textbf{BLEU-3}} \\ \hline
{\ul \textbf{S2S Baselines}}    &                                      &                                      &                                     &                                      \\
BASE                            & 23.15                                & 21.62                                & 20.87                               & 2.67                                 \\
 COPY                          & 32.58                                & 30.61                                & 25.82                               & 5.58                                 \\
TOPIC                         & 33.60                                & 31.59                                & 28.55                               & 8.46                                 \\ \hline
{\ul \textbf{Current SOTA}}     &                                      &                                      &                                     &                                      \\
DUAL DEC                        & 34.98                                & 32.84                                & 29.41                               & 8.84                                 \\
T5                              & 45.33                                & 42.69                                & 37.34                               & 21.06                                \\ 
\lyx{ChatGLM3}                             & 44.54                                & 41.87                                & 34.31                               & 18.34                                \\ 
\hl{GPT-3.5-turbo}                        & 32.41                                & 28.48                                & 24.91                               & 5.74                                 \\ \hline
{\ul \textbf{Our Model}}        &                                      &                                      &                                     &                                      \\
UniPoll                         & \textbf{47.92}                       & \textbf{45.02}                       & \textbf{39.96}                      & \textbf{22.78}                       \\ \hline
\end{tabular}
}
\end{table}
\begin{table}
\centering
\caption{\hl{Comparison results on \textit{RedditPolls} for poll generation. All baselines are fine-tuned, except GPT-3.5-turbo prompted in a one-shot manner. UniPoll significantly outperforms all other models (paired t-test; p-value \textless 0.05).}}
\label{tab:reddit_polls_PG}
\resizebox{1\columnwidth}{!}
{
\begin{tabular}{|l|rrrr|}
\hline
\multirow{2}{*}{\textbf{Model}} & \multicolumn{4}{c|}{\textbf{Poll Generation}}                                                                                                            \\
                                & \multicolumn{1}{c}{\textbf{ROUGE-1}} & \multicolumn{1}{c}{\textbf{ROUGE-L}} & \multicolumn{1}{c}{\textbf{BLEU-1}} & \multicolumn{1}{c|}{\textbf{BLEU-3}} \\ \hline
{\ul \textbf{S2S Baselines}}    &                                      &                                      &                                     &                                      \\
BASE                            & 9.79                                & 8.10                                & 5.22                               & 1.31                                 \\ \hline
{\ul \textbf{Current SOTA}}     &                                      &                                      &                                     &                                      \\
T5                              & 38.50                                & 36.80                                & 27.15                               & 16.52                                \\
ChatGLM3                              & 36.55                                & 34.38                                & 28.16                               & 16.14                                \\ 
GPT-3.5-turbo                              & 32.01                                & 27.71                                & 18.69                               & 6.81                                \\\hline
{\ul \textbf{Our Model}}        &                                      &                                      &                                     &                                      \\
UniPoll                         & \textbf{43.64}                       & \textbf{41.69}                       & \textbf{32.04}                      & \textbf{20.46}                       \\ \hline
\end{tabular}
}
\end{table}

\subsection{Main Comparisons}\label{subsection:overall_performance}

Here we first measure the overall performance of poll generation with automatic and human evaluations.

\paragraph{Automatic Evaluations}
The results on standard benchmark \textit{WeiboPolls} are shown in Table \ref{tab:main_results_PG}, from which we can draw the following observations.

First, the performance gain achieved by DUAL DEC and TOPIC is somewhat limited compared to COPY.
It suggests that while incorporating topic models and dual decoder design, as proposed in previous work \cite{lu_engage_2021}, can partly address the challenges, its overall benefits to the task may be constrained.

Second, by integrating pre-trained NLG, \lyx{ChatGLM3 and T5 demonstrate outstanding performance compared to DUAL DEC, improving the results by large margins.
It can be attributed to using a pre-trained NLG model (with decoder and (or) encoder pre-trained), which acquires generic reading and writing skills through large-scale pre-training.}
Also, considering prior findings \cite{lu_engage_2021} implying the limited assistance provided by pre-trained encoders (such as BERT \cite{devlin_bert_2018-1}), we can infer that a pre-trained decoder is more crucial in social media poll generation.


Third, UniPoll significantly outperforms all baselines and previous state-of-the-art models. Specifically, UniPoll achieves a ROUGE-1 score of 47.92, exceeding the performance of both T5 (45.33) and DUAL DEC (34.98). \hl{When compared with large language models such as GPT-3.5-turbo and ChatGLM3, UniPoll also demonstrates superior results.} This improvement can be attributed to the multi-objective optimization design, which enables the model to better capture the nuanced relationships between context, questions, and answers. \hl{Notably, GPT-3.5-turbo, as a proprietary large language model, suffers from performance limitations when relying solely on prompting techniques, further illustrating the advantage of our lightweight approach.}
In addition to improved performance, UniPoll produces more stable outputs, with a mean standard deviation of 0.25, reflecting an 81.48\% reduction compared to T5. This underscores the positive influence of multi-objective optimization on both output quality and consistency.

\hl{A similar conclusion can be drawn from the curated English benchmark \textit{RedditPolls}, as shown in Table }\ref{tab:reddit_polls_PG}
\hl{, further demonstrating the generalization capability of UniPoll across different languages and cultural contexts. Given the comparable results, our subsequent experiments will be conducted on the standard \textit{WeiboPolls} benchmark unless otherwise specified.}



\paragraph{Human Ratings}
\begin{table}
\centering
\caption{Human Evaluation Results on \textit{WeibiPolls} for Poll Generation. Higher scores indicate better results.}
\label{tab:hunman_evaluation}
\resizebox{1\columnwidth}{!}
{
\begin{tabular}{|l|r|r|r|}
\hline
\textbf{Model} & \textbf{Relevance} & \textbf{Fluency} & \textbf{Engagingess} \\ \hline
Gold Standard  & 2.79               & 2.84             & 2.74                 \\ \hline
BASE           & 1.26               & 2.14             & 1.35                 \\
TOPIC        & 1.81               & 1.66             & 1.50                 \\
DUAL DEC       & 2.02               & 1.87             & 1.67                 \\
T5             & 2.34               & 2.28             & 2.07                 \\
\lyx{ChatGLM3}     & 2.31               & 2.37             & 2.01                 \\
\hl{GPT-3.5-turbo}     & 1.97               & 1.92             & 1.62                 \\
UniPoll        & \textbf{2.46}      & \textbf{2.44}    & \textbf{2.20}        \\ \hline
\end{tabular}
}
\end{table}
To further examine the usability of generated polls, we followed the benchmark setup \cite{lu_engage_2021} and enlisted the assistance of four native Chinese speakers to rate the generated results manually.
Our evaluation criteria included three metrics: 1) \textit{Relevance}, measuring the poll's relevance to the source posts;
2) \textit{Fluency}, assessing the readability and flow of the generated language;
3) \textit{Engagingness}, gauging the degree of attractiveness of the polls in drawing engagement.
Each metric adopts a 4-point Likert scale, with scores of 1 for \textit{very bad}, 2 for \textit{bad}, 3 for \textit{good}, and 4 for \textit{very good}.
We selected the first 100 sample outputs from the test set for human rating, and the raters were unaware of which model the results came from for an unbiased examination.
Table \ref{tab:hunman_evaluation} displays the average ratings from the four annotators.

\begin{table}
\centering
\vspace{-0.35cm}
\caption{Question Generation Comparison on \textit{WeiboPolls}. 
}
\label{tab:main_results_QG}
\resizebox{1\columnwidth}{!}
{
\begin{tabular}{|l|rrrr|}
\hline
\multirow{2}{*}{\textbf{Model}} & \multicolumn{4}{c|}{\textbf{Question Generation}}                                                                                                        \\
                                & \multicolumn{1}{c}{\textbf{ROUGE-1}} & \multicolumn{1}{c}{\textbf{ROUGE-L}} & \multicolumn{1}{c}{\textbf{BLEU-1}} & \multicolumn{1}{c|}{\textbf{BLEU-3}} \\ \hline
{\ul \textbf{S2S Baselines}}    &                                      &                                      &                                     &                                      \\
BASE                            & 21.62                                & 20.64                                & 20.35                               & 2.11                                 \\
COPY                          & 35.13                                & 33.20                                & 30.27                               & 7.95                                 \\
TOPIC                         & 36.65                                & 34.70                                & 31.11                               & 8.66                                 \\ \hline
{\ul \textbf{Current SOTA}}     &                                      &                                      &                                     &                                      \\
DUAL DEC                        & 38.24                                & 36.14                                & 32.27                               & 9.04                                 \\
T5                              & 44.46                                & 42.06                                & 36.91                               & 16.26                                \\
\lyx{ChatGLM3}                        & 45.78                                & 42.93                                & 33.26                               & 13.37                                \\
\hl{GPT-3.5-turbo}                       & 32.83                                & 29.27                                & 26.73                               & 5.21                                 \\ \hline
{\ul \textbf{Our Model}}        &                                      &                                      &                                     &                                      \\
UniPoll                         & \textbf{49.60}                       & \textbf{46.71}                       & \textbf{42.04}                      & \textbf{19.83}                       \\ \hline
\end{tabular}
}
\end{table}
\begin{table}[t]
    \centering
    \vspace{-0.35cm}
    \caption{Answer Generation Comparison on \textit{WeiboPolls}. 
    }
    \label{tab:main_results_AG}
    \resizebox{1\columnwidth}{!}
    {
    \begin{tabular}{|l|rrrr|}
    \hline
    \multirow{2}{*}{\textbf{Model}} & \multicolumn{4}{c|}{\textbf{Answers Generation}}                                                                                                         \\
                                    & \multicolumn{1}{c}{\textbf{ROUGE-1}} & \multicolumn{1}{c}{\textbf{ROUGE-L}} & \multicolumn{1}{c}{\textbf{BLEU-1}} & \multicolumn{1}{c|}{\textbf{BLEU-3}} \\ \hline
    {\ul \textbf{S2S Baselines}}    &                                      &                                      &                                     &                                      \\
    BASE                            & 24.68                                & 22.59                                & 21.38                               & 3.22                                 \\
    COPY                          & 30.03                                & 28.02                                & 21.38                               & 3.22                                 \\
    TOPIC                         & 30.56                                & 28.49                                & 26.00                               & 8.26                                 \\ \hline
    {\ul \textbf{Current SOTA}}     &                                      &                                      &                                     &                                      \\
    DUAL DEC                        & 31.72                                & 29.54                                & 26.55                               & 8.65                                 \\
    T5                              & 46.20                                & 43.32                                & 37.77                               & \textbf{25.86}                       \\
    \lyx{ChatGLM3}                        & 43.29                                & 40.81                                & 35.35                               & 23.21                                \\
    \hl{GPT-3.5-turbo}                        & 32.00                                & 27.70                                & 23.09                               & 6.27                                 \\ \hline
    {\ul \textbf{Our Model}}        &                                      &                                      &                                     &                                      \\
    UniPoll                         & \textbf{46.24}                       & \textbf{43.34}                       & \textbf{37.87}                      & 25.74                                \\ \hline
    \end{tabular}
    }
    \end{table}

The results show that human-generated polls, represented by the Gold Standard, perform the best in all three evaluation metrics.
This suggests that machine-generated polls still have some room for improvement.
Among the machine-generated models, UniPoll achieves the highest scores in all three metrics, indicating its effectiveness in generating human-favored social media polls, as the multi-objective optimization helps balance well on context, question, and answer learning.
The DUAL DEC model and TOPIC outperform the BASE model in terms of relevance and engagement; however, they lag in fluency, potentially due to the side effects of incorporating latent topics into the decoders. \lyx{Meanwhile, ChatGLM3 demonstrates better fluency compared to T5, which can be partially attributed to the vast amount of pretraining data and the extensive model parameters}
Nevertheless, with a pre-trained decoder, T5, ChatGLM3 and UniPoll, outperform BASE on fluency, showing the helpfulness of NLG pre-training on our task.

In addition, as existing benchmark  \cite{lu_engage_2021} only examines the overall poll quality, we further probe into the internal structure of a poll and assess its question-answer consistency.
T5 and UniPoll received scores of 1.96 and 2.50, respectively, indicating that UniPoll's multi-objective optimization strategy can helpfully capture question-answer relations.


\subsection{Separate Evaluation on Question and Answer Generation }\label{subsection:individual_performance}

We have demonstrated UniPoll’s effectiveness in overall poll generation. Here, we provide a detailed comparison of model performance in question and answer generation on \textit{WeiboPolls}, as shown in Table \ref{tab:main_results_QG} and Table \ref{tab:main_results_AG}. \hl{Similar conclusions can be drawn from the results on \textit{RedditPolls} in Table} \ref{tab:reddit_polls_QG} \hl{and Table }\ref{tab:reddit_polls_AG}.

Comparing question and answer generation results, UniPoll exhibits more significant improvement in assisting question formation.
\lyx{For example, it achieves a ROUGE-1 score of 49.60, 5.14 and 3.82 points higher than T5 and ChatGLM3.}
This result indicates that UniPoll's multi-objective optimization strategy enables the model to effectively learn across different tasks and better extract the focus of the discussion from the context to generate high-quality questions.
In the answer generation task, T5's performance is competitive to UniPoll because it has two decoders dedicated to learning question and answer generation separately.
As a result, it may somehow overfit one objective (answer generation), sacrificing the other (question generation) and the question-answer consistency (as shown in Section \ref{subsection:overall_performance}).
Nevertheless, UniPoll coordinates well with question-and-answer generation to result in better poll generation.

\begin{table}
\centering
\vspace{-0.35cm}
\caption{\hl{Question Generation Comparison on \textit{RedditPolls}. 
}}
\label{tab:reddit_polls_QG}
\resizebox{1\columnwidth}{!}
{
\begin{tabular}{|l|rrrr|}
\hline
\multirow{2}{*}{\textbf{Model}} & \multicolumn{4}{c|}{\textbf{Poll Generation}}                                                                                                            \\
                                & \multicolumn{1}{c}{\textbf{ROUGE-1}} & \multicolumn{1}{c}{\textbf{ROUGE-L}} & \multicolumn{1}{c}{\textbf{BLEU-1}} & \multicolumn{1}{c|}{\textbf{BLEU-3}} \\ \hline
{\ul \textbf{S2S Baselines}}    &                                      &                                      &                                     &                                      \\
BASE                            & 10.54                                & 8.75                               & 5.64                               & 1.48                                 \\ \hline
{\ul \textbf{Current SOTA}}     &                                      &                                      &                                     &                                      \\
T5                              & 33.13                                & 31.32                                & 23.67                               & 12.70                                \\
ChatGLM3                              & 32.64                                & 30.16                                & 28.45                               & 13.62                                \\ 
GPT-3.5-turbo                              & 23.28                                & 21.86                                & 8.62                               & 3.96                                \\\hline
{\ul \textbf{Our Model}}        &                                      &                                      &                                     &                                      \\
UniPoll                         & \textbf{39.57}                       & \textbf{37.32}                       & \textbf{29.92}                      & \textbf{16.97}                       \\ \hline
\end{tabular}
}
\end{table}
\begin{table}
\centering
\vspace{-0.35cm}
\caption{\hl{Answer Generation Comparison on \textit{RedditPolls}. 
}}
\label{tab:reddit_polls_AG}
\resizebox{1\columnwidth}{!}
{
\begin{tabular}{|l|rrrr|}
\hline
\multirow{2}{*}{\textbf{Model}} & \multicolumn{4}{c|}{\textbf{Poll Generation}}                                                                                                            \\
                                & \multicolumn{1}{c}{\textbf{ROUGE-1}} & \multicolumn{1}{c}{\textbf{ROUGE-L}} & \multicolumn{1}{c}{\textbf{BLEU-1}} & \multicolumn{1}{c|}{\textbf{BLEU-3}} \\ \hline
{\ul \textbf{S2S Baselines}}    &                                      &                                      &                                     &                                      \\
BASE                            &  9.04                                & 7.44                                & 4.79                                & 1.13                                 \\ \hline
{\ul \textbf{Current SOTA}}     &                                      &                                      &                                     &                                      \\
T5                              & 43.86                                & 42.28                                & 30.62                               & 20.34                                \\
ChatGLM3                              & 40.46                                & 38.59                                & 27.86                               & 18.65                                \\ 
GPT-3.5-turbo                              & 40.73                                & 33.56                                 & 28.76                                & 9.66                                \\\hline
{\ul \textbf{Our Model}}        &                                      &                                      &                                     &                                      \\
UniPoll                         & \textbf{47.71}                       & \textbf{46.05}                       & \textbf{34.15}                      & \textbf{23.94}                       \\ \hline
\end{tabular}
}
\end{table}

\subsection{Automated Context Enrichment}\label{subsection:context_aug}

\hl{User comments play a crucial role in enhancing the quality of generated content (as further discussed in Section }\ref{subsection:ablations}\hl{. However, comments often contain irrelevant or noisy information, such as advertisements or spam. Moreover, in real-time poll generation for commercial platforms, comments may be unavailable at the time a post is made, making it essential to explore alternative methods for enriching context.}
\hl{This raises an important question: can we augment the context in the absence of user comments? In this section, we explore the effectiveness of Retrieval-Augmented Generation (RAG) }\cite{gao2024retrievalaugmented}\hl{ and synthetic comment generation techniques as a means to enhance context understanding when comments are noisy or unavailable.}

\begin{table}
\centering
\caption{\lyx{ROUGE-1 Score for Different Methods without Comments. UniPoll+RAG and UniPoll+Synthetic are two augmented methods that employ Retrieval-Augmented Generation and synthetic comment techniques.}}
\label{tab:rag_synthetic}
\resizebox{1\columnwidth}{!}
{
\begin{tabular}{|l|ccc|}
\hline
\multirow{2}{*}{\textbf{Method}} & \textbf{Poll}       & \textbf{Question}   & \textbf{Answers}    \\
                                 & \textbf{Generation} & \textbf{Generation} & \textbf{Generation} \\ \hline
T5                               & 39.43               & 39.47               & 39.40               \\
UniPoll                          & 41.48               & 43.96               & 39.01               \\
UniPoll+Synthetic                & 42.16               & 44.12               & 40.20               \\
UniPoll+RAG                      & \textbf{42.39}      & \textbf{44.43}      & \textbf{40.35}      \\ \hline
\end{tabular}
}
\end{table}

\lyx{UniPoll+RAG employs Retrieval-Augmented Generation (RAG) }
\cite{gao2024retrievalaugmented}
\lyx{ techniques by integrating posts retrieved as supplementary material, enabling the model to enrich its context without relying on direct user comments. }
\lyx{Specifically, we utilized a Weibo post database as the source of information and employed FlagEmbedding } 
\cite{bge_embedding} 
\lyx{ for vectorization. Due to memory constraints, we incorporated the top five relevant posts as supplementary resources, as demonstrated in Table }
\ref{tab:rag_synthetic}
\lyx{, UniPoll+RAG reaches peak performance of 42.39 in poll generation with 0\% real comments, marking a clear enhancement over the baseline UniPoll model.}
\lyx{ More importantly, we observe a significant improvement in answer generation, with an increase of 1.34 for ROUGE-1 scores, demonstrating the model's crucial ability to extract diverse opinions.}

\lyx{Alternatively, UniPoll+Synthetic employs a different strategy by using the large language model ChatGLM3 }\cite{du2022glm}\lyx{ to generate synthetic comments. This approach aims to simulate user interactions and generate artificial feedback that can act as a proxy for real comments. By integrating these synthetic comments, the model tests the hypothesis that synthetic data can replace real user interactions in training scenarios. As shown in Table }\ref{tab:rag_synthetic}\lyx{, this method shows promise, although it does not achieve the same performance peaks as UniPoll+RAG.}

Through these methodologies, we address and mitigate noisy, insufficient comments, particularly in early stages of content creation. Both RAG and synthetic approaches represent promising avenues for enhancing the model's context understanding, underscoring the potential of these techniques to improve content generation in environments with limited user interaction. Further refinement of these strategies is necessary to fully capitalize on their benefits.

\subsection{Ablation Study}\label{subsection:ablations}
\begin{table}
\centering
\caption{Abaltion Test of UniPoll.}
\label{tab:ablations}
\resizebox{1\columnwidth}{!}
{
\begin{tabular}{|l|rrrr|}
\hline
\textbf{Model}                     & \textbf{ROUGE-1} & \textbf{ROUGE-L} & \textbf{BLEU-1} & \textbf{BLEU-3} \\ \hline
{\ul \textbf{Poll Generation}}     &                  &                  &                 &                 \\
UniPoll                            & \textbf{47.92}   & \textbf{45.02}   & \textbf{39.96}  & \textbf{22.78}  \\
w.o. A                             & 47.60            & 44.87            & 39.58           & 22.70           \\
w.o. Q                             & 47.81            & 44.92            & 39.84           & 22.69           \\
w.o. Q, A                          & 47.68            & 44.80            & 39.86           & 22.75           \\ \hline
{\ul \textbf{Question Generation}} &                  &                  &                 &                 \\
UniPoll                            & 49.60            & 46.71            & 42.04           & 19.83           \\
w.o. A                             & \textbf{49.70}   & \textbf{46.81}   & \textbf{42.10}  & \textbf{20.00}  \\
w.o. Q                             & 49.18            & 46.29            & 41.62           & 19.41           \\
w.o. Q, A                          & 49.43            & 46.61            & 42.02           & 19.99           \\ \hline
{\ul \textbf{Answers Generation}}  &                  &                  &                 &                 \\
UniPoll                            & 46.24            & 43.34            & 37.87           & 25.74           \\
w.o. A                             & 45.49            & 42.92            & 37.07           & 25.41           \\
w.o. Q                             & \textbf{46.45}   & \textbf{43.54}   & \textbf{38.07}  & \textbf{25.97}  \\
w.o. Q, A                          & 45.93            & 42.99            & 37.70           & 25.52           \\ \hline
\end{tabular}
}
\end{table}

The previous discussions have shown the positive effects of multi-objective optimization, adopting multi-task learning with two auxiliary tasks in a unified framework.
While it helps UniPoll outperform the previous state-of-the-art models, we are interested in the relative contributions of different auxiliary tasks and hence present an ablation study in Table \ref{tab:ablations}.

As can be seen, the simultaneous inclusion of two auxiliary tasks (UniPoll) resulted in the greatest performance improvements for the poll generation task.
Compared to the other ablations, the highest scores were obtained when training with auxiliary tasks for generating questions (Q) and answers (A) only.
This indicates the effectiveness of our multi-objective optimization strategy.
For the poll generation, different auxiliary tasks show various impacts, with task A having a relatively larger impact than Q.
The reason could be that generating answers is inherently more challenging than questions because answers, comprising a multitude of potential choices, usually present scattered elements, whereas a question, in sentence form, has easier-to-catch syntax by pre-trained NLG models.
For this reason, incorporating task A appears to help the model strengthen the learning for answer generation.
Our findings are consistent with the previous QAG work \cite{wang_joint_2017}, which suggests that answer generation needs more help than question generation in the joint training framework.

We also find ablation w.o Q performs the best for question generation while w.o A champions answer generation.
These indicate that adding one auxiliary task may negatively affect the other auxiliary task, suggesting the challenges to balancing Q and A objectives in our main task (poll generation).
Nevertheless, UniPoll effectively coordinates the two objectives and outputs the best poll generation results, showing the helpfulness of adopting prompt tuning for task distinguishing and multi-objective optimization for task collaboration.

\lyx{Further testing the model under different weight ratios confirms these observations. As shown in Table }\ref{tab:ablations_ratio}\lyx{, the equal weight ratio (1,1) for both $\gamma_Q$ (question generation weight) and $\gamma_A$  (answer generation weight) achieves the highest scores across all metrics, indicating a balanced contribution of both tasks to the overall performance of the model. Adjusting the weights reveals slight performance variations, yet the model consistently shows optimal results with a balanced weight setting. This robustness across various configurations suggests that while UniPoll can function effectively under a range of weight settings, a balanced approach best optimizes relevance and fluency, thereby enhancing the quality of the generated polls. The detailed examination of these weight variations not only underscores the model's flexibility but also emphasizes the critical role of finely tuned task weighting in achieving high-quality NLG outputs in social media contexts. This analysis affirms the effectiveness of our multi-objective optimization strategy and underscores the importance of appropriate task weighting in complex NLG tasks like poll generation.}

\begin{table}
\centering
\caption{\lyx{Abaltion Test of Different Weight Ratios on Poll Generation.}}
\label{tab:ablations_ratio}
\resizebox{1\columnwidth}{!}
{
\begin{tabular}{|cc|rrrr|}
\hline
\textbf{$\gamma_Q$} & \textbf{$\gamma_A$} & \textbf{ROUGE-1} & \textbf{ROUGE-L} & \textbf{BLEU-1} & \textbf{BLEU-3} \\ \hline
1   & 1   & \textbf{47.92} & \textbf{45.02} & \textbf{39.96} & \textbf{22.78} \\
0.8 & 1   & 47.35 & 44.61 & 39.35 & 22.53 \\
0.6 & 1   & 47.49 & 44.75 & 39.49 & 22.62 \\
0.4 & 1   & 47.36 & 44.66 & 39.37 & 22.46 \\
0.2 & 1   & 47.03 & 44.34 & 39.01 & 22.29 \\
1   & 0.8 & 47.34 & 44.60 & 39.41 & 22.53 \\
1   & 0.6 & 47.35 & 44.67 & 39.40 & 22.53 \\
1   & 0.4 & 47.22 & 44.54 & 39.27 & 22.45 \\
1   & 0.2 & 47.06 & 44.40 & 38.99 & 22.20 \\ \hline
\end{tabular}
}
\end{table}

\subsection{Quantitative Analysis}\label{subsection:quantitative-analysis}

\begin{figure*}[!t]
	\centering
	\subfloat[Poll Generation]{
		\label{fig:Weibo-Few-QA}
		\includegraphics[width=0.32\linewidth]{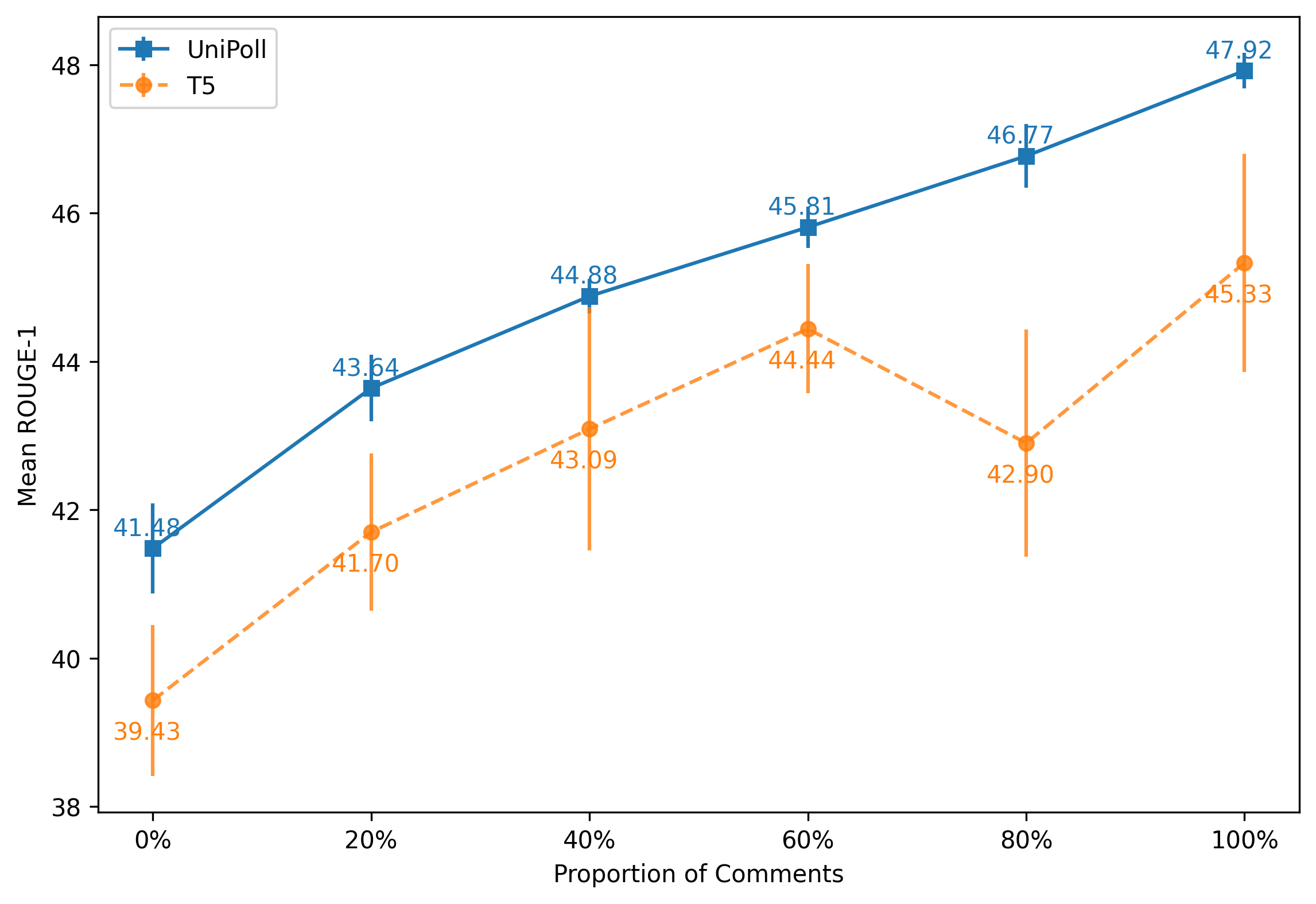}}
    \hfil
	\subfloat[Question Generation]{
		\label{fig:Weibo-Few-Q}
		\includegraphics[width=0.32\linewidth]{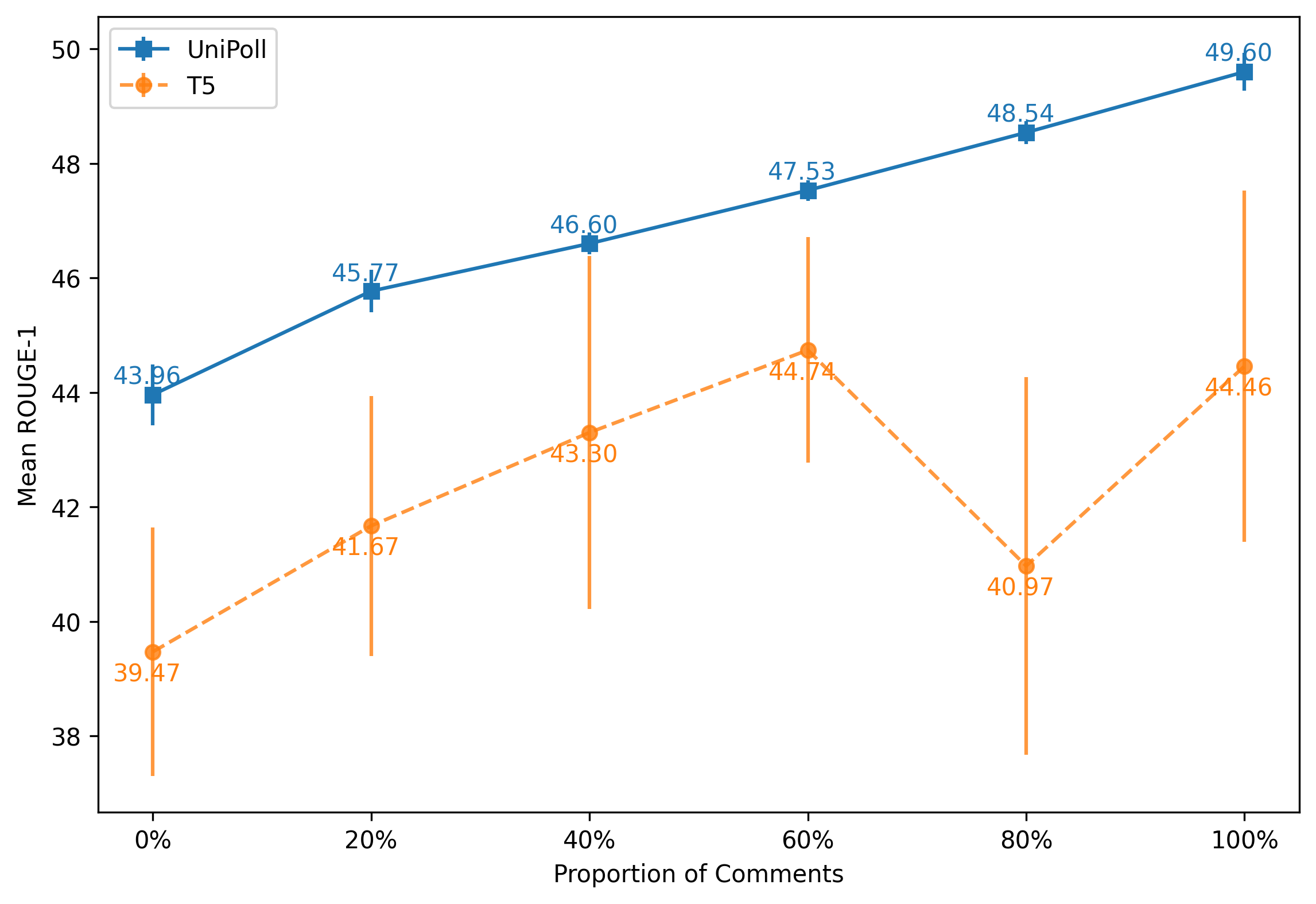}}
    \hfil
	\subfloat[Answers Generation]{
		\label{fig:Weibo-Few-A}
		\includegraphics[width=0.32\linewidth]{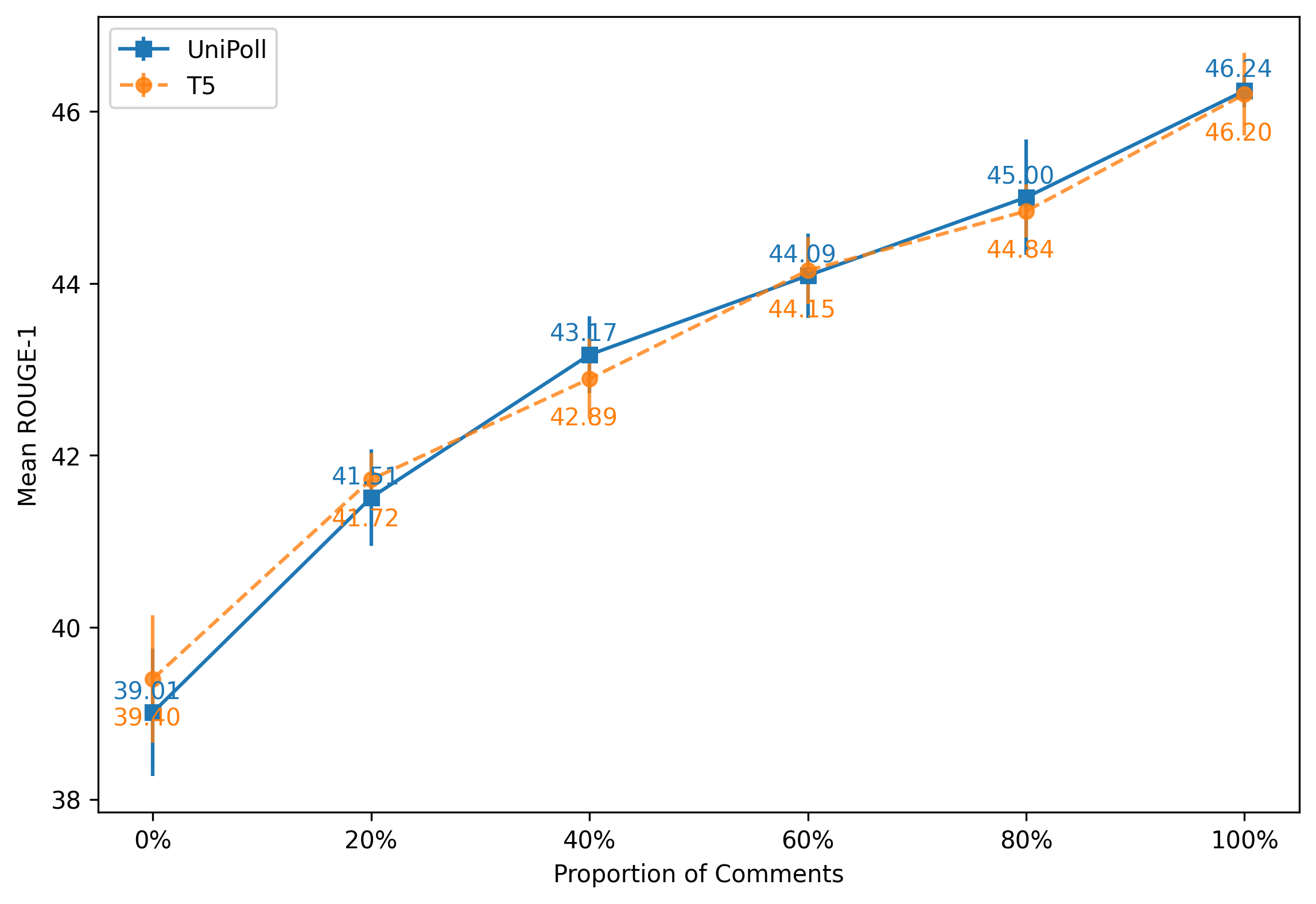}}
	\caption{\lyx{Generation results from models trained with varying scales of comment data on \textit{WeiboPolls}. The X-axis represents the percentage of comments used for training, while the Y-axis displays the ROUGE-1 scores. The blue solid lines depict the performance of UniPoll, and the orange dashed lines represent T5.}}
	\label{fig:Weibo-Few}
\end{figure*}

The previous experiments are conducted on the entire dataset.
Readers may then be curious about models' sensitivity to varying training samples whose input contexts involve posts and comments.
Hence, we first discuss the effects of comments on training poll generation, followed by training data scales.

\paragraph{Effects of Comments}

Here we quantify the results of UniPoll trained with varying early proportions of comments.
Thus, comments were sorted chronologically, and the first $n$\% of comments were employed for training and testing.
Figure \ref{fig:Weibo-Few} depicts UniPoll and T5's ROUGE-1 scores with varying $n$.

Our results demonstrate that including comments substantially improves the quality of generated polls, resulting in a ROUGE-1 score improvement of over 6 points across different tasks and models.
This finding is consistent with previous work \cite{lu_engage_2021}, showing the helpfulness of comments in enriching the contexts for short posts.
We also observe that UniPoll performs consistently better than T5, given varying scales of comments.
It is because UniPoll allows a better understanding of context-question-answer relations through multi-objective optimization with the two auxiliary tasks.
The gain is obvious when there are no comments (0\%), where UniPoll outperforms T5 on question generation with a ROUGE-1 score of 44.96, which is 5.49 higher than T5's score of 39.47.

Moreover, interestingly, T5's performance is extremely unstable when a large number of comments are provided, possibly because the noise in comments may mislead the model in extracting relevant context.
In contrast, UniPoll exhibits stronger stability in all three tasks, showing its superiority in handling noisy data with a progressive performance improvement as the percentage of comments increases.



\begin{figure*}[!t]
	\centering
	\subfloat[Poll Generation]{
		\label{fig:Weibo-Low-QA}
		\includegraphics[width=0.32\linewidth]{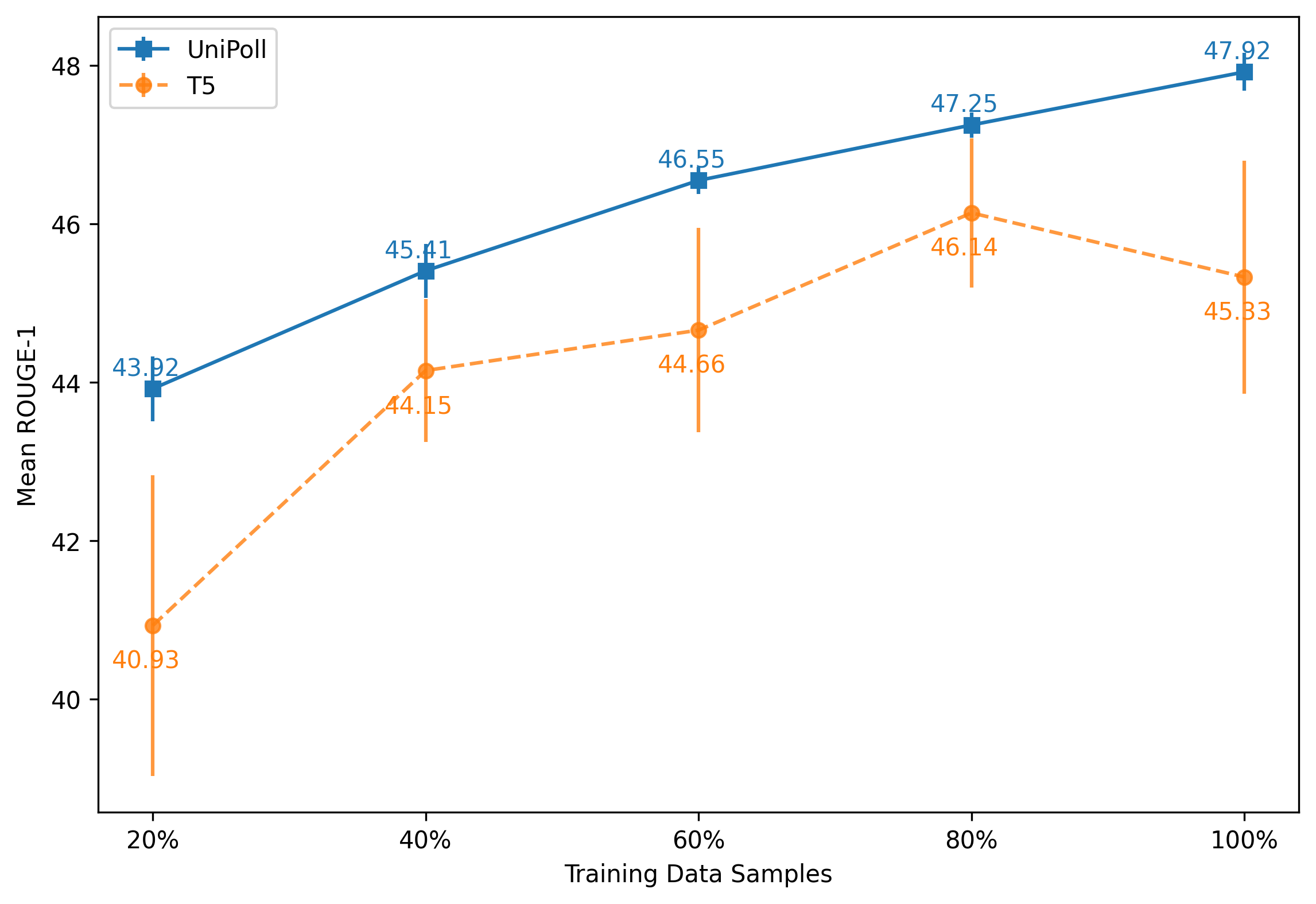}}
    \hfil
	\subfloat[Question Generation]{
		\label{fig:Weibo-Low-Q}
		\includegraphics[width=0.32\linewidth]{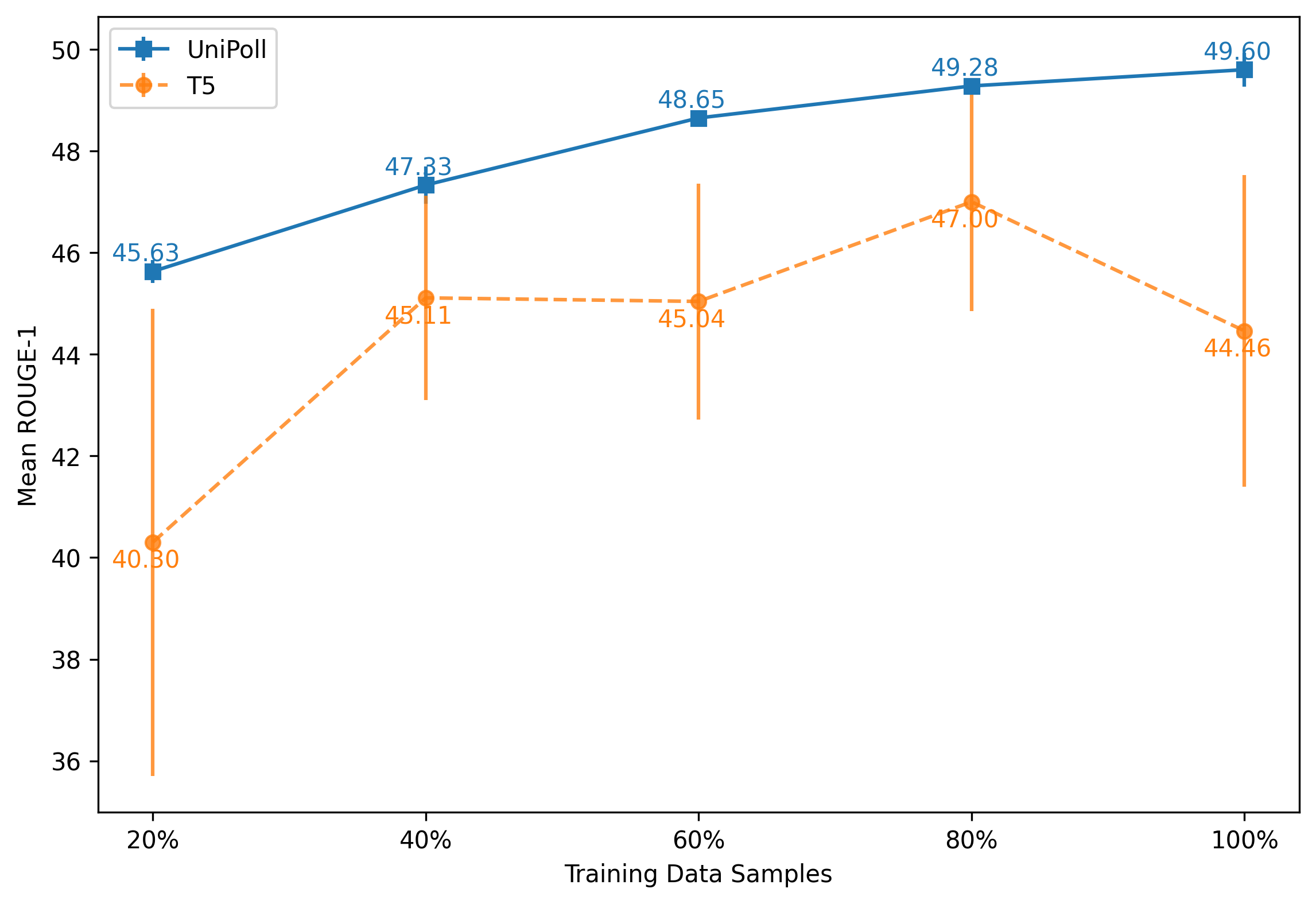}}
    \hfil
	\subfloat[Answers Generation]{
		\label{fig:Weibo-Low-A}
		\includegraphics[width=0.32\linewidth]{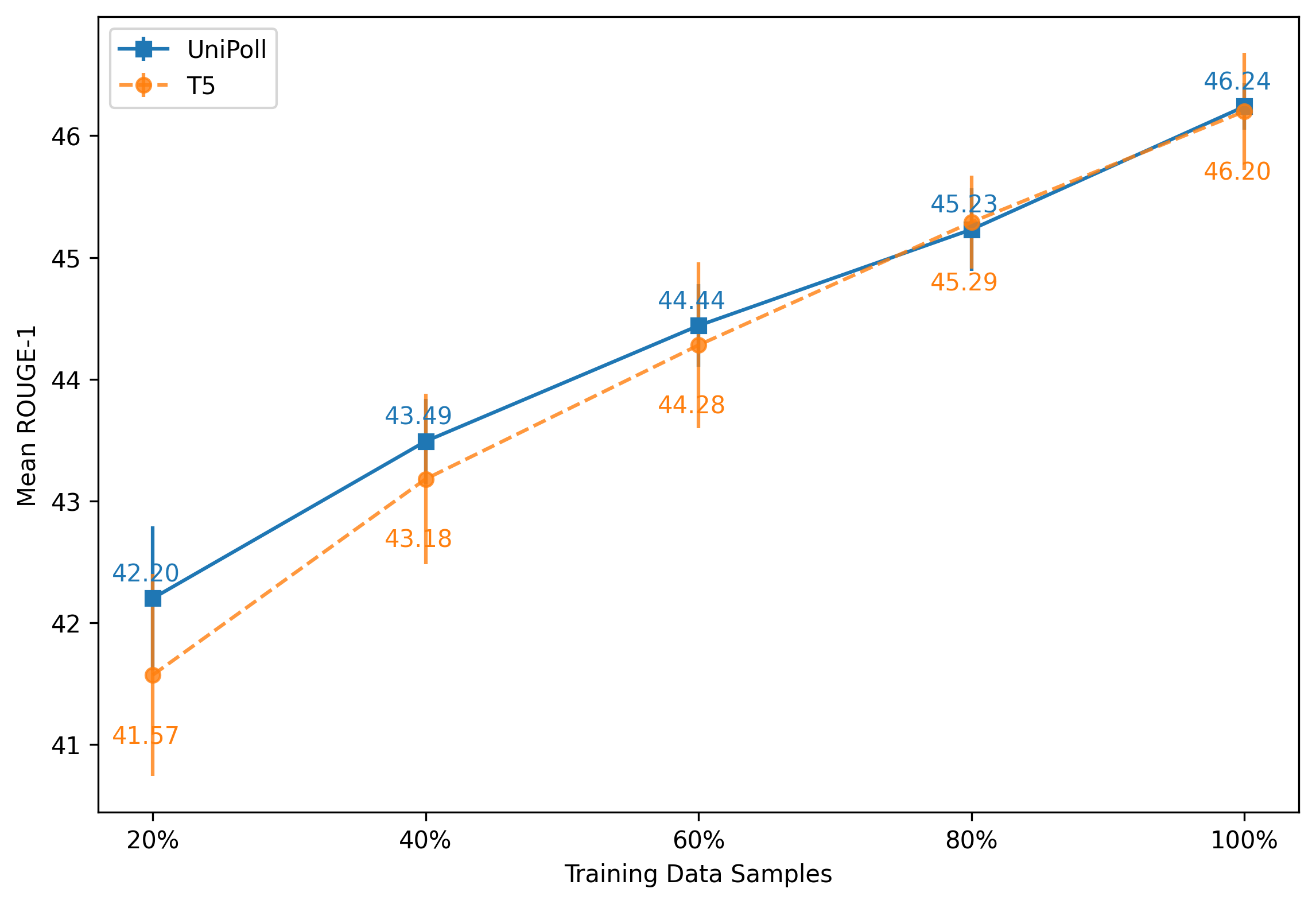}}
	\caption{\lyx{Generation results of models trained with varying scales of training data from \textit{WeiboPolls}. X-axis shows the percentage of training data from 20\% to 100\%, and the y-axis indicates the ROUGE-1 scores. Blue solid lines indicate UniPoll's results, and orange dotted lines are T5's.}}
	\label{fig:low_resource}
\end{figure*}

\paragraph{Effects of Training Data Scales}
We then study the results when the models are trained with varying data scales to assess model effectiveness in low-resource situations.
The models were trained using varying amounts of data - 20\%, 40\%, 60\%, 80\%, and 100\%.
Figure \ref{fig:low_resource} shows the results.

UniPoll performs consistently better in varying training data scales for the poll and question generation.
A large performance gain is shown when the models are trained with extremely limited data (20\%).
The two models exhibit similar results for answer generation, whereas UniPoll's results are slightly better with smaller training data scales.
These results indicate UniPoll's multi-objective optimization increases its data-using efficiency, leading to superiority in low-resource training where data is scarce or costly.

\subsection{Case Study}\label{subsection:case_study}

\begin{figure*}[!h]
	\centering
	\subfloat[Case 1]{
		\begin{tcolorbox}[width=0.48\linewidth, fontupper=\footnotesize, fontlower=\footnotesize, colback=white, boxrule=1pt, arc=0mm, equal height group=A]
            [\textit{\textbf{Post}}]: \#线上复试是否能保障公平\# 高考延期惹的祸，考研线上复试，那还能保证公平吗？(\#Can online retesting ensure fairness\# The disruptions caused by the postponement of the entrance exams have resulted in postgraduate retests being conducted online. Can this method still ensure fairness?)\par\vspace{0.1cm}

            [\textit{\textbf{Comments}}]: 这个世界上本来就没有绝对的公平。你可以说一个倒数第一考了第一，但考上了他也还是啥都不会。也可以说他会利用一切机会达到目的，反正结果就是人家考的好，你还找不出来证据。线上考试，平时考倒数的人进了年级前十。平时考试有水分，线上之后，那不就是在水里考？ (Absolute fairness doesn't exist in this world. Even if someone who ranks last comes first in their exams, it doesn't mean they possess the necessary knowledge when they enter university. They will seize every opportunity to achieve their goals, resulting in good exam performance, without any tangible evidence against them. In the case of online exams, individuals who typically rank last may end up in the top ten of their grade. Regular exams can be subject to manipulation, and when conducted online, there may be even more room for unforeseen circumstances.)\par

            \tcbline 
            [\textit{\textbf{Gold}}]: 你觉得线上复试是否公平 (Do you think the online retest is fair)\par
            \textgreater 公平 (Fair); 不公平 (Unfair); 其他 (Others) \par\vspace{0.1cm}

            [\textit{\textbf{T5}}]: 你支持线上复试吗 (Do you support online retesting)\par
            \textgreater 能 (Can); 不能 (Cannot) \par\vspace{0.1cm}

            [\textit{\textbf{UniPoll}}]: 你认为线上复试公平吗 (Do you think the online retest is fair)\par
            \textgreater 公平 (Fair); 不公平 (Unfair); 评论区补充 (Please comment if you have more to say.) \par\vspace{0.1cm}

            [\textit{\textbf{ChatGPT}}]: 线上考试公平性投票 (Online exam fairness poll)\par
            \textgreater 保障公平 (Guaranteeing fairness); 存在瑕疵 (Deficiencies exist); 无法确定 (Not sure) \par\vspace{0.1cm}

            [\textit{\textbf{NewBing}}]: 线上复试是否能保障公平 (Can online retesting guarantee fairness)\par
            \textgreater 可以 (Yes); 不可以 (No); 不确定 (Not sure) \par

            \tcbline 
            [\textit{\textbf{T5 w.o. comments}}]: 你觉得线上复试能保障公平吗 (Do you think online retesting guarantees fairness)\par
            \textgreater 能 (Yes); 不能 (No); 看情况 (Depends) \par\vspace{0.1cm}

            [\textit{\textbf{UniPoll w.o. comments}}]: 你认为线上复试公平吗 (Do you think the online retest is fair)\par
            \textgreater 公平 (Fair); 不公平 (Unfair); 看情况 (Depends) \par\vspace{0.1cm}

            [\textit{\textbf{ChatGPT w.o. comments}}]: 考研线上复试公平性投票 (Online retest fairness poll)\par\vspace{0.1cm}
            \textgreater 能保证公平 (Fairness guaranteed);不能保证公平 (Fairness can't be guaranteed); 取消线上复试 (Eliminate online retesting) \par\vspace{0.1cm}

            [\textit{\textbf{NewBing w.o. comments}}]: Failed to generate\par
            \textgreater Failed to generate \par
        \end{tcolorbox}
        \label{fig:case1}
        }
	\subfloat[Case 2]{
		\begin{tcolorbox}[width=0.48\linewidth, fontupper=\footnotesize, fontlower=\footnotesize, colback=white, boxrule=1pt, arc=0mm, equal height group=A]
            [\textit{\textbf{Post}}]: \#哪吒，大鱼海棠重映\# 动画电影《哪吒之魔童降世》、《大鱼海棠》，以及雷佳音、佟丽娅主演的《超时空同居》确定将重映。据最新数据显示，3月24日全国复工影院495家，复工率4.36\%，单日票房2.7万元。 (\#Nezha, Big Fish and Begonia to be Re-released\# The animated films "Nezha: Birth of the Demon Child" and "Big Fish and Begonia," along with "How Long Will I Love U," starring Lei Jiayin and Tong Liya, have been confirmed for re-release. According to the latest data, on March 24, a total of 495 cinemas nationwide resumed operations, with a resumption rate of 4.36\% and a single-day box office of 27,000 yuan.)\par\vspace{0.1cm}

            [\textit{\textbf{Comments}}]: 我在人间贩卖黄昏，只为收集世间温柔，去见你。谢谢你的分享，来看看你。我的微博，随时恭候你的到来... (I am selling Twilight on Earth, just to collect the tenderness of the world to meet you. Thank you for sharing and coming to see you. My Weibo feeds, waiting for you anytime...)\par

            \tcbline 
            [\textit{\textbf{Gold}}]: 你最期待哪部重新上映 (Which one are you most looking forward to re-releasing)\par
            \textgreater 哪吒之魔童降世 (Nezha: Birth of the Demon Child); 大鱼海棠 (Big Fish and Begonia); 超时空同居 (How Long Will I Love U) \par\vspace{0.1cm}

            [\textit{\textbf{T5}}]: 你会去看哪部 (Which one would you go to watch)\par
            \textgreater 期待 (Expectations); 不期待 (No expectations)\par\vspace{0.1cm}

            [\textit{\textbf{UniPoll}}]: 你最期待哪部电影重映 (Which film are you most looking forward to re-releasing)\par
            \textgreater 哪吒之魔童降世 (Nezha: Birth of the Demon Child); 超时空同居 (How Long Will I Love U); 都期待 (Both); 都不期待 (None)\par\vspace{0.1cm}

            [\textit{\textbf{ChatGPT}}]: 你更想看哪部电影 (Which film would you prefer to watch)\par
            \textgreater 哪吒之魔童降世 (Nezha: Birth of the Demon Child); 大鱼海棠 (Big Fish and Begonia); 超时空同居 (How Long Will I Love U)\par\vspace{0.1cm}

            [\textit{\textbf{NewBing}}]: 哪吒大鱼海棠重映，你会选择哪部电影 (Which film would you choose in a re-release of Nezha or Big Fish and Begonia?)\par
            \textgreater  哪吒之魔童降世 (Nezha: Birth of the Demon Child); 大鱼海棠 (Big Fish and Begonia); 都想看 (Both)\par

            \tcbline 
            [\textit{\textbf{T5 w.o. comments}}]: 你觉得哪部电影最值得一看 (Which film do you think is most worth watching)\par
            \textgreater 期待 (Expectations); 不期待 (No expectations) \par\vspace{0.1cm}

            [\textit{\textbf{UniPoll w.o. comments}}]: 你还会去看重映的电影吗 (Do you still go to re-releases)\par
            \textgreater 会 (Yes); 不会 (No); 看情况 (Depends); \par\vspace{0.1cm}

            [\textit{\textbf{ChatGPT w.o. comments}}]: 选择你要观看的电影吧 (Select the film you want to watch)\par\vspace{0.1cm}
            \textgreater 哪吒之魔童降世 (Nezha: Birth of the Demon Child); 大鱼海棠 (Big Fish and Begonia); 超时空同居 (How Long Will I Love U)\par\vspace{0.1cm}

            [\textit{\textbf{NewBing w.o. comments}}]: Failed to generate\par
            \textgreater Failed to generate\par
        \end{tcolorbox}
        \label{fig:case2}
        }
	\caption{Case studies about online retesting for the national postgraduate entrance exam (Case 1), and the first post-COVID-19 movie to watch (Case 2). Each case is displayed in three parts. The top presents the input to the model, including the [\textit{Post}] and [\textit{Comments}]. In the middle, we show the model-generated polls, [\textit{T5}] and [\textit{UniPoll}], with the author-written gold standard, [\textit{Gold}]. Besides, we include evaluations of the latest models, [\textit{ChatGPT}][\textit{NewBing}].
 At the bottom, we show the model results without taking comments for training (e.g. [\textit{UniPoll w.o. comments}]) to examine the qualitative effects of comments.
 }
    \label{fig:case_study}
\end{figure*}

We have experimented UniPoll in a quantitative view.
Here we use two cases as examples to qualitatively discuss model output and show the results in Figure \ref{fig:case_study}.
Specifically, we involve the cases from \textit{ChatGPT}\cite{openai_gpt-4_2023} (March-23 version) and \textit{NewBing}\cite{mehdi_reinventing_2023}, which are unable to be included in the main comparison because they require case-sensitive prompts to guide the generation.
For reproduction concerns, the prompts we adopt in the case studies are shown in the Appendix.

In Case 1, UniPoll demonstrates a high-quality poll whose generated question adeptly captures the input post's specific concerns (about exam fairness) and well reflects that point in the question.
Moreover, all three answer choices show high consistency with the poll question.
On the contrary, T5 suffers from inconsistency between the question (``Do you...") and answers (``Can or Cannot") because the QA relations were not carefully exploited.
ChatGPT, without learning from social media data, managed to generate a fine poll yet somehow looks too serious, lacking the potential to engage social media engagements.
NewBing's results look good with sufficient information, whereas it fails to generate a poll without comments even though we tried multiple times with varying prompts.


Case 2 exemplifies a scenario with implicit context-question-answer relationships and noisy social media data, where comments discuss topics unrelated to the post and provide useless contexts.
Despite the challenging conditions, UniPoll successfully captures the point about ``movie re-releasing'' from the post and generates a poll closely related to it.
In contrast, T5 again suffers from QA inconsistency problems and lacks details in the output.

In summary of the case study results, we observe UniPoll's outputs are usually specifically related to the context's key points, whereas others' outputs are relatively more general.
It is attributed to UniPoll's efforts to strengthen learning in social media context-question-answer relations.
For this reason, its output polls are also more engaging, which is helpful in drawing other users' attention to participating in discussions.

\subsection{Error Analysis}\label{subsection:error_analysis}

The above results have demonstrated the potential of UniPoll in handling poll generation.
To provide a more comprehensive view and insight into its limitations, we analyze several errors that UniPoll made in 100 sampled cases.

\paragraph{Error Analysis for Question Generation}
We first investigate question generation and discuss the error types.

\textbf{(i) Alternative Questions.}
Topic diversity poses a significant challenge in question generation (12/100), occurring when the context contains multiple possible topics.
Consequently, the model generates a question alternative to the gold standard.
For instance, in a post discussing smelly foods like durian and stinky tofu, the gold question was ``你认为哪个最臭" (``Which do you think is the smelliest?”), while UniPoll generated ``你最喜欢哪一道" (``Which dish do you like the most?”).
Due to the open and informal social media writing styles, such phenomena are common.
Although UniPoll may generate an appropriate question, the evaluation could label it as an error due to dissimilarity with the ground truth, revealing the limitations of current evaluation metrics and the need for better NLG testing methods.

\textbf{(ii) Grammar Errors.}
A small percentage of generated questions (4/100) contain grammar errors, such as ``你会安利别人安利自己吗" (``Do you enthuse others enthuse yourself?") or ``婚后不想和父母同住吗” (``Do you not want to live with your parents after marriage?”).
These errors often stem from noisy training data.
Although minor, such grammatical issues can affect user experience and the perceived quality of polls.
Future work could incorporate cleaner training data and post-processing checks to ensure grammatical accuracy.

\textbf{(iii) Off-Topic Questions.}
While UniPoll generally demonstrated good context understanding, a few instances (2/100) contained off-topic questions.
This issue may arise from noisy input contexts, complicating topic identification.
However, due to the small error sample size, deeper analysis is limited.

\paragraph{Error Analysis for Answer Generation}

For answer generation, the errors fall into the following types:

\textbf{(i) Duplicated Choices.}
In 9/100 cases, generated answers contained duplicates.
Although correct and complete, such duplication may bias polling results.
Post-processing steps could mitigate this error.

\textbf{(ii) QA Inconsistency.}
Despite UniPoll’s improvements over T5, QA inconsistency persists in 8/100 cases, leaving room for further enhancement.
This issue likely stems from the limited training data for question-answer consistency.
Future work may employ data augmentation to expand training data and address this problem.

\textbf{(iii) Information Insufficiency.}
In 2/100 cases, errors stemmed from insufficient context information.
Analysis of UniPoll variants without comment training revealed 10/100 such errors, indicating comments help enrich context.
Future efforts could enhance models' common-sense reasoning and leverage external knowledge for better context encoding.

        \section{Conclusion}\label{section:conclusion}

In this work, we addressed the challenge of automatically generating social media polls, comprising an open-ended question and multiple answer choices, based on the context of a social media post and its comments. This task is uniquely challenging for existing NLG models due to the informal, noisy nature of social media text and implicit relationships between context, question, and answers.
We introduced UniPoll, a novel multi-objective optimization framework built on the pre-trained T5 model, designed to effectively learn context-poll, context-question, and context-answer relations. By leveraging enriched context from user comments and incorporating Retrieval-Augmented Generation (RAG) and synthetic data generation, UniPoll overcomes the challenges of noisy social media data.
Experimental results on large-scale \textit{WeiboPolls} and the newly introduced \textit{RedditPolls} datasets demonstrate UniPoll’s state-of-the-art performance, significantly surpassing models like T5, ChatGLM3, and GPT-3.5. Quantitative analysis highlights UniPoll’s robustness in low-resource training, while qualitative evaluations confirm it produces more contextually relevant and engaging polls than other models. UniPoll offers a powerful tool to enhance user interaction and engagement on social media platforms.

\section{Limitations and Future Work}

\lyx{While this study has advanced our understanding of poll generation using natural language processing, it also illuminates several areas where enhancements could be beneficial. This section outlines current limitations and proposes future research directions to address these challenges and expand the scope of our findings.}

\begin{enumerate}
    \item \textbf{\lyx{Performance in Few Comments Scenarios:}}
    \lyx{As shown in Section }\ref{section:results}\lyx{, Retrieval-Augmented Generation and synthetic techniques significantly improved performance in scenarios with few comments. Future work could refine these methods by integrating real-time data from related posts or trending topics to further enhance poll generation.}

    \item \textbf{\lyx{Multilingual and Cross-Cultural Expansion:}}
    \lyx{Adapting UniPoll for diverse linguistic and cultural contexts requires more than translation; it demands accounting for cultural nuances influencing user interactions. Developing a culturally aware model with diverse global datasets would enable broader applicability.}

    \item \textbf{\lyx{Integration of Multimodal Data:}}
    \lyx{With social media increasingly sharing text, images, videos, and links in posts, future UniPoll versions could enhance relevance and engagement by incorporating multimodal inputs. Techniques from computer vision and audio processing could help analyze the sentiment and content of images and videos, informing the poll generation process.}

    \item \textbf{\lyx{Ethical and Responsible AI Use:}}
    \lyx{Adhering to ethical AI practices is essential. Future developments should establish guidelines to prevent reinforcing stereotypes or spreading misinformation, ensuring responsible content generation.}

    \end{enumerate}

\lyx{Addressing these challenges and exploring these research directions will enhance UniPoll’s responsiveness, cultural sensitivity, and ethical standards. These efforts will prepare UniPoll to better serve the diverse needs of global social media landscapes, making it a more effective tool for engaging users and capturing public sentiment.}
        
\ifCLASSOPTIONcompsoc
  \section*{Acknowledgments}
\else
  \section*{Acknowledgment}
\fi

We would like to acknowledge the use of artificial intelligence (AI)-generated text in this paper. Specifically, we utilized the \textit{ChatGPT} and \textit{NewBing} language models developed by OpenAI and Microsoft for our case study. The sections of this paper that contain AI-generated text have been duly cited, referencing the respective AI system used. We are grateful to the creators and developers of \textit{ChatGPT} and \textit{NewBing} for providing these powerful language models, which contributed to the research presented in this paper.

%

{\appendix[Prompt Template for ChatGPT and NewBing]

\begin{center}
\begin{tcolorbox}[width=0.98\linewidth, fontupper=\footnotesize, fontlower=\footnotesize, colback=white, boxrule=1pt, arc=0mm]
    [\textit{\textbf{Case 1}}]: 
    要求：根据以下微博内容和微博评论，生成一个简短的投票标题和几个选项，要求选项的字数不超过10个字，标题字数不超过20个字。
    微博内容：线上复试是否能保障公平高考延期惹的祸考研线上复试那还能保证公平吗。
    微博评论：这个世界上本来就没有绝对的公平你可以说一个倒数第一考了第一但考上了他也还是啥都不会也可以说他会利用一切机会达到目的反正结果就是人家考的好你还找不出来证据线上考试平时考倒数的人进了年级前十好平时考试有水分线上之后那不就是在水里考。\par
                
    \tcbline 
    
    [\textit{\textbf{Case 1 w.o. comments}}]: 
    要求：根据以下微博内容，生成一个简短的投票标题和几个选项，要求选项的字数不超过10个字，标题字数不超过20个字。
    微博内容：线上复试是否能保障公平高考延期惹的祸考研线上复试那还能保证公平吗。\par
    
    \tcbline 
    
    [\textit{\textbf{Case 2}}]: 
    要求：根据以下微博内容和微博评论，生成一个简短的投票标题和几个选项，要求选项的字数不超过10个字，标题字数不超过20个字。
    微博内容：哪吒大鱼海棠重映动画电影哪吒之魔童降世大鱼海棠以及雷佳音佟丽娅主演的超时空同居确定将重映据最新数据显示DIGIT月DIGIT日全国复工影院DIGIT家复工率DIGIT单日票房DIGIT万元。
    微博评论：我在人间贩卖黄昏只为收集世间温柔去见你谢谢你的分享来看看你我的微博随时恭候你的到来一个今天胜过两个明天一个评论胜过两个今天中午好我来啦支持不变微博在于互动所以我来了眼泪不是答案拼搏才是选择只有回不了的过去没有到不了的明天看电影可以找我啊绝对精彩你要热门了老铁没有人能一路单纯到底但别忘了最初的自己好久不见啦我自带小板凳来的。\par
                
    \tcbline 
    
    [\textit{\textbf{Case 2 w.o. comments}}]: 
    要求：根据以下微博内容，生成一个简短的投票标题和几个选项，要求选项的字数不超过10个字，标题字数不超过20个字。
    微博内容：哪吒大鱼海棠重映动画电影哪吒之魔童降世大鱼海棠以及雷佳音佟丽娅主演的超时空同居确定将重映据最新数据显示DIGIT月DIGIT日全国复工影院DIGIT家复工率DIGIT单日票房DIGIT万元。\par
\end{tcolorbox}
\end{center}

}

\ifCLASSOPTIONcaptionsoff
  \newpage
\fi



\bibliographystyle{IEEEtran}
\bibliography{IEEEabrv,references.bib}

%


    \end{CJK}
\end{document}